\documentclass{svmono}
\usepackage{mathptmx,helvet,courier,type1cm} 
\usepackage{graphicx,url,amsmath,amssymb,dsfont}
\graphicspath{{figures}}
\begin{document}

\chapauthor{Jos\'e Lezama, Samy Blusseau,
            Jean-Michel Morel, Gregory Randall, Rafael Grompone von Gioi}
\chapter{Psychophysics, Gestalts and Games}

\begin{quote}\textbf{Abstract:}
Many psychophysical studies are dedicated to the evaluation of the human gestalt
detection on dot or Gabor patterns, and to model its dependence on the pattern
and background parameters. Nevertheless, even for these constrained percepts,
psychophysics have not yet reached the challenging prediction stage, where human
detection would be quantitatively predicted by a (generic) model. On the other
hand, Computer Vision has attempted at defining automatic detection
thresholds. This chapter sketches a procedure to confront these two
methodologies inspired in gestaltism.

Using a computational quantitative version of the non-accidentalness principle,
we raise the possibility that the psychophysical and the (older) gestaltist
setups, both applicable on dot or Gabor patterns, find a useful complement in a
Turing test. In our perceptual Turing test, human performance is compared by the
scientist to the detection result given by a computer. This confrontation
permits to revive the abandoned method of gestaltic games. We sketch the
elaboration of such a game, where the subjects of the experiment are confronted
to an alignment detection algorithm, and are invited to draw examples that will
fool it. We show that in that way a more precise definition of the alignment
gestalt and of its computational formulation seems to emerge.

Detection algorithms might also be relevant to more classic psychophysical
setups, where they can again play the role of a Turing test. To a visual
experiment where subjects were invited to detect alignments in Gabor patterns,
we associated a single function measuring the alignment detectability in the
form of a number of false alarms (NFA). The first results indicate that the
values of the NFA, as a function of all simulation parameters, are highly
correlated to the human detection. This fact, that we intend to support by
further experiments, might end up confirming that human alignment detection is
the result of a single mechanism.
\end{quote}

\section{Introduction}

Alan Turing advanced a controversial proposal in 1950 that is now known as the
\emph{Turing Test} \cite{turing1950}. Turing's aim was to discuss the problem of
machine intelligence and, instead of giving a premature definition of thinking,
he framed the problem in what he called the \emph{Imitation Game}: A human
interrogator interacts with another human and a machine, but only in typewritten
form; the task of the interrogator is to ask questions in order to determine
which of its two interlocutors is the human. Turing proposed that a machine that
eventually could not be distinguished from humans by its answers should be
considered intelligent. This influential suggestion sparked a fruitful debate
that continues to this day \cite{pinar2000turing}.

Our concern here is however slightly different. We are studying perception and
Turing precluded in his test any machine interaction with the environment other
that the communication through the teletype; he concentrated on the pure problem
of thinking and to that aim avoided fancy computer interactions, that anyway did
not exist at his time. Yet, machine perception is still a hard problem for which
current solutions are far from the capacities of humans or animals\footnote{It
  is a common practice in Internet services to use the so-called CAPTCHAs to
  ensure that the interaction is made with a human and not an automatic
  program. A CAPTCHA, which stands for \emph{Completely Automated Public Turing
    test to tell Computers and Humans Apart}, usually consists in a perceptual
  task, simple to perform for humans but hard for known algorithms. This
  suggests that visual and auditive perception currently provides the most
  effective Turing test.}. Our purpose is to discuss a variety of
\emph{perceptual imitation games } as a research methodology to develop machine
vision algorithms on the one hand, and quantitative psychophysical protocols on
the other.

Human perceptual behavior has been the subject of quantitative experimentation
since the times of Fechner, the founder of Psychophysics. This relatively new
science investigates the relationship between the stimulus intensity and the
perceived sensation \cite{stevens}. But this approach does not provide a
perceptual theory in which machine vision and an imitation game could be based.

The Gestalt school, Wertheimer, K\"ohler, Koffka, Kanizsa among others
\cite{wertheimer, kohler, source-book, metzger, kanizsa}, developed from the
twenties to the eighties an original \emph{modus operandi}, based on the
invention and display to subjects of clever geometric figures
\cite{Wagemans12CenturyOfGestalt-I, Wageman12CenturyOfgestalt-II}. A
considerable mass of experimental evidence was gathered, leading to the
conclusion that the first steps of visual perception are based on a reduced set
of geometrical grouping laws. Unfortunately these Gestalt laws, relevant though
they were, remained mainly qualitative and led to no direct machine perception
approach.

Since the emergence of the field of Computer Vision \cite{marr} about fifty
years ago -- initially as a branch of the Artificial Intelligence working with
robots and its artificial senses -- there have been many attempts at formalizing
vision theories and especially Gestalt theory \cite{sarkar1993}. Among them one
finds models of neural mechanism \cite{grossberg1985neural}, theories based on
logical inference \cite{feldman1997regularity}, on information theory
\cite{leclerc1989constructing}, invoking minimum description principles
\cite{zhu1996region}, or grammars of visual elements
\cite{zhu-mumford,han-zhu}. Nevertheless, only a small fraction of these
proposals has been accompanied by systematic efforts to compare machine and
human vision. An important exception is the Bayesian theory of perception
\cite{mumford} that has attracted considerable attention in cognitive sciences,
leading to several experimental evaluations \cite{Feldman01BayesianContour,
  Kersten04ObjectPerceptionBayesianInference}. A recent groundbreaking work by
Fleuret~et~al. \cite{geman-etal-pnas} compared human and machine performing
visual categorization tasks. Humans are matched against learning algorithms in
the task of distinguishing two classes of synthetic patterns. One class for
example may contain four parallel identical shapes in arbitrary position, while
the other class contains the same shapes but with arbitrary orientation and
position. It was observed that humans learn the distinction of such classes with
very few examples, while learning algorithms require considerably more examples,
and nevertheless gain a much lower classification performance. The experimental
design was more directed at pointing out a flaw of learning theory, though, than
at contributing to psychophysics.

Such experiments stress the relevance of computer vision as a research program
in vision, in addition to a purely technological pursuit. Its role should be
complementary to explanatory sciences of natural vision by providing, not only
descriptive laws, but actual implementations of mechanisms of operation. With
that aim, perceptual versions of the imitation game should be the Leitmotiv in
the field, guiding the conception, evaluation and success of theories.

Here we will present comparisons of human perception to algorithms based on the
\emph{non-accidentalness principle} introduced by Witkin, Tenenbaum and Lowe
\cite{Lowe85,witkin1,witkin2} as a general grouping law. This principle states
that spatial relations are perceptually relevant only when their accidental
occurrence is unlikely. We shall use the \emph{a contrario} framework, a
particular formalization of the principle due to Desolneux, Moisan and Morel
\cite{DMM2003,DMM_book} as part of an attempt to provide a mathematical
foundation to Gestalt Theory.

This chapter is intended to give an overview of our research program; for this
reason we reduced the settings to the bare minimum, concentrating in one simple
geometric structure, namely alignments. The methodology however is general. By
using such a simple structure we will present two complementary aspects of the
same program, each one with specific imitation games: a research procedure
inspired in the methodology of the Gestalt school and the use of online games
for psychophysical experimentation.

Gestaltism created clever figures in which humans fail to perceive the expected
structures, generating illusions. In the \emph{gestaltic game}, as we shall call
our first proposed methodology, the experimenter tries to fool the algorithm by
building a particular data set that produces unnatural results. This methodology
is discussed in Sect.~\ref{sec:gestalt}, along with a brief introduction to the
\emph{a contrario} methods.

The second part, in Sect.~\ref{sec:psychophysics}, is dedicated to a first
attempt at a psychophysical evaluation of the same theory. There is a difference
with classic psychophysical experiments in which detection thresholds are
measured; here each stimulus will be shown to human subjects but also to an
algorithm, and both will answer yes or no to the visibility of a given
structure. In a second variation, both humans and machine will also have to
point to the position of the observed structure. This last variation is proposed
as an online game, used as a methodology to facilitate experimentation and the
attraction of volunteers.

Being the result of a work in progress, no final conclusion will be drawn. Our
overall goal is to advocate for new sorts of quantitative Gestalt and
psychophysical games.

\section{Detection Theory versus Gestaltism}\label{sec:gestalt}

Here and in most of the text we shall call ``gestalt'' any geometric structure
emerging perceptually against the background in an image. We stick to this
technical term because it is somewhat untranslatable, meaning something between
``form'' and ``structure''. According to Gestalt theory, the gestalts emerge by
a grouping process in which the properties of similarity (by color, shape,
texture, etc.), proximity, good continuation, convexity, parallelism, alignment
can individually or collaboratively stir up the grouping of the building
elements sharing one or more properties.

\subsection{The Gestaltic Game}

One of the procedures used by Gestalt psychology practitioners was to create
clever geometric figures that would reveal a particular aspect of perception
when used in controlled experiments with human subjects. They pointed out the
grouping mechanisms, but also the striking fact that geometric structures
objectively present in the figure are not necessarily part of the final gestalt
interpretation. These figures are in fact counterexamples against simplistic
perception mechanisms. Each one represents a challenge to a theory of vision
that should be able to cope with all of them.

The methodology we propose in order to design and improve automatic geometric
gestalt detectors is in a way similar to that of the gestaltist. One starts with
a primitive method that works correctly in very simple examples. The task is
then to produce data sets where \emph{humans} clearly see a particular gestalt
while the rudimentary method produces a different interpretation. Analyzing the
errors of the first method gives hints to improve the procedure in order to
create a second one that produces better results with the whole data set
produced until that point. The same procedure is applied to the second method to
produce a third one, and successive iterations refine the methods step by
step. The methodology used by the Gestalt psychologist to study human perception
is used here to push algorithms to be similar to their natural
counterpart. Finding counterexamples is less and less trivial after some
iterations and the counter-examples become, like gestaltic figures, more and
more clever.

We decided to render this process interactive by drawing figures in a computer
interface that delivers a detection result immediately. The exploration of
counterexamples is in that way transformed into an active search where previous
examples are gradually modified in an attempt to fool the detection
algorithm. The figures are all collected to be later used at the analysis
stage. The \emph{gestaltic game} is at the same time a method to produce
interesting data sets, a methodology to develop new detection algorithms and a
collaborative tool for research in the computational gestalt community. Each
detection game will only stop when it eventually passes the Turing test, the
algorithm's detection capability becoming undistinguishable from that of a
human.

\subsection{Dot Alignments Detection}

For its simplicity, dot patterns are often used in the study of visual
perception. Several psychophysical studies led by Uttal have investigated the
effect of direction, quantity and spacing in dot alignment perception
\cite{Uttal70,Uttal73}. The detection of collinear dots in noise was the target
of a study attempting to assess quantitatively the masking effect of the
background noise \cite{Tripathy99}. A recent work by Preiss analyzes various
perceptual tasks on dot patterns from a psychophysical and computational
perspective \cite{PreissThesis}. An interesting computational approach to detect
gestalts in dot patterns is presented in \cite{Ahuja89}, although the study is
limited to very regularly sampled patterns. A practical application of alignment
detection is presented in \cite{Vanegas10}.

\begin{figure}[b]
\centering
\includegraphics[width=1\linewidth]{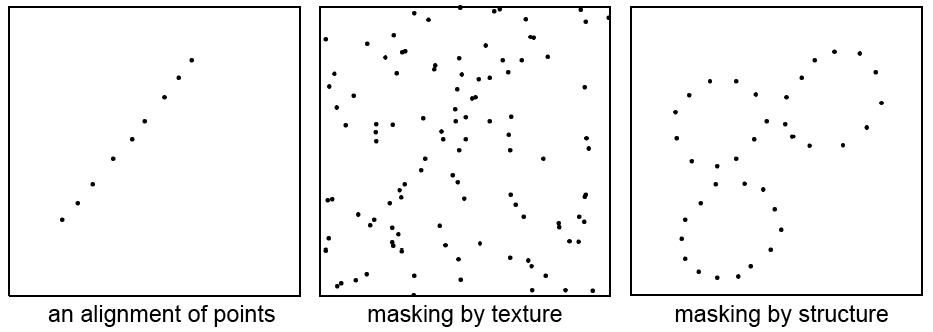}
\caption{Exactly the same set of aligned dots is present in the three images,
  but it is only perceived as such in the first one. The second one is a classic
  masking by texture case and the third a masking by structure one, often called
  ``Gestalt conflict''.}
\label{fig:masking_examples}
\end{figure}

From a gestaltic point of view, a point alignment is a group of points sharing
the property of being aligned in one direction. While it may seem a simple
gestalt, Fig.~\ref{fig:masking_examples} shows how complex the alignment event
is. From a purely factual point of view, the same alignment is present in the
three figures. However, it is only perceived as such by most viewers in the
first one. The second and the third figures illustrate two occurrences of the
\emph{masking phenomenon} discovered by gestaltists \cite{kanizsa:vedere}: the
\emph{masking by texture}, which occurs when a gestalt is surrounded by a
clutter of randomly distributed similar objects or \emph{distractors}, and the
\emph{masking by structure}, which happens when the alignment is masked by other
perceptually more relevant gestalts, a phenomenon also called \emph{perceptual
  conflict} by gestaltists \cite{metzger,metzger:en,kanizsa}. The magic
disappearance of the alignment in the second and third figures can be accounted
for in two very different ways. As for the first one, we shall see that a
probabilistic \emph{a contrario} model \cite{DMM_book} is relevant and can lead
to a quantitative prediction. As for the second disappearance, it requires the
intervention of another more powerful grouping law, the \emph{good continuation}
\cite{kanizsa1979organization}.

These examples show that a mathematical definition of dot alignments is required
before even starting to discuss how to detect them. A purely geometric-physical
description is clearly not sufficient to account for the masking
phenomenon. Indeed, an objective observer making use of a ruler would be able to
state the existence of the very same alignment at the same precision on all
three figures. But this statement would contradict our perception, as it would
contradict any reasonable computational (definition and) theory of alignment
detection.

This experiment also shows that the detection of an alignment is highly
dependent on the context of the alignment. It is therefore a complex question,
and must be decided by building mathematical definitions and detection
algorithms, and confronting them to perception. As the patterns of
Fig.~\ref{fig:masking_examples} already suggest, simple computational
definitions with increasing complexity will nevertheless find perceptual
counterexamples. There is no better way to describe the ensuing ``computational
gestaltic game'' than describing how the dialogue of more and more sophisticated
alignment detection algorithms and counterexamples help build up a perception
theory.

\subsection{Basic Dot Alignment Detector}\label{sec:basic_dot_align}

A very basic idea that could provide a quantitative context-dependent definition
of dot alignments is to think of them as thin, rectangular shaped point
clusters. In that case, the key measurements would be the relative dot densities
inside and outside the rectangle. The algorithm described in this section
follows the \emph{a contrario} methodology \cite[Sect.~3.2]{DMM_book} according
to which a group of elements is detectable as a gestalt if and only if it has a
low enough probability of occurring just by chance in an \emph{a contrario}
background model.

We shall first introduce briefly the \emph{a contrario} framework
\cite{DMM_book,dmm2000,DMM2003}. The approach is based on the
\emph{non-accidentalness principle}
\cite{witkin1,wagemans-nonaccidental,es2003computational,spelke1990principles}
(sometimes called \emph{Helmholtz principle}) that states that structures are
perceptually relevant only when they are unlikely to arise by accident. An
alternative statement is ``we do not perceive any structure in a uniform random
image'' \cite[p.31]{DMM_book}. The \emph{a contrario} framework is a particular
formalization of this principle adjusting the detection thresholds so that the
\emph{expected} number of accidental detections is provably bounded by a small
constant $\varepsilon$. The key point is how to define \emph{accidental}
detections. This requires a stochastic model, the so-called \emph{a contrario}
model, characterizing unstructured or random data in which the sought gestalt
could only be observed by chance.

\begin{figure}[t]
\centering
\fbox{\includegraphics[width=.4\linewidth]{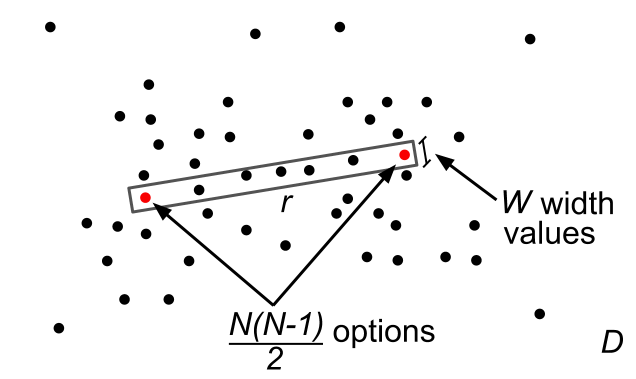}}
\caption{A schematic representation of the evaluated rectangle. In an image with
  $N$ points, there are $(N(N-1)\times W)/2$ possible rectangles defined by two
  dots. In the case shown in this figure, $N = 49$ and $k(r,\mathbf{x}) = 5$.}
\label{fig:basic_algorithm_schema}
\end{figure}

Consider a dot pattern defined on a domain $D$ with total area $S_D$ and
containing $N$ dots, see Fig.~\ref{fig:basic_algorithm_schema}. We are
interested in detecting groups of dots that are well aligned. A first reasonable
\emph{a contrario} hypothesis $H_0$ for this problem is to suppose that the $N$
dots are the result of a random process where points are independent and
uniformly distributed in the domain. The question is then to evaluate whether
the presence of aligned points contradicts the \emph{a contrario} model or not.

Given an observed set of $N$ points $\mathbf{x}=\{x_i\}_{i=1\ldots N}$ and a
rectangle $r$ (the candidate to contain an alignment), we will denote by
$k(r,\mathbf{x})$ the number of those points observed inside $r$. We decide
whether to keep this candidate or not based on two principles: a good candidate
should be non-accidental, and any equivalent or better candidate should be kept
as well. The degree of non-accidentalness of an observed rectangle $r$ can be
measured by how small the probability $\mathds{P}\big[k(r,\mathbf{X})\geq
  k(r,\mathbf{x})\big]$ is, where $\mathbf{X}$ denotes a random set of $N$ dots
following $H_0$. In the same vein, a rectangle $r^{\prime}$ will be considered
at least as good as $r$ given the observation $\mathbf{x}$, if
$\mathds{P}\big[k(r^{\prime},\mathbf{X})\geq k(r^{\prime},\mathbf{x})\big] \leq
\mathds{P}\big[k(r,\mathbf{X})\geq k(r,\mathbf{x})\big]$.

Recall that we want to bound the expected number of accidental detections.
Given that $N_{tests}$ candidates will be tested, the expected number of
rectangles which are as good as $r$ under $H_0$, is about \cite{DMM_book}
\begin{equation}\label{eq:expected_number_of_events}
   N_{tests} \cdot\mathds{P}\Big[k(r,\mathbf{X})\geq k(r,\mathbf{x})\Big].
\end{equation}
The $H_0$ stochastic model fixes the probability law of the random number of
points in the rectangle, $k(r,\mathbf{X})$, which only depends on the total
number of dots $N$. The discrete nature of this law implies that
(\ref{eq:expected_number_of_events}) is not actually the expected value but an
upper bound of it \cite{DMM_book, GJ08}. Let us now analyze the two factors in
(\ref{eq:expected_number_of_events}).

Here the \emph{a contrario} model $H_0$ assumes that the $N$ points are
i.i.d. with uniform density on the domain. Under the \emph{a contrario}
hypothesis $H_0$, the probability that one dot falls into the rectangle $r$ is
\begin{equation}
   p = \frac{S_r}{S_D},
\end{equation}
where $S_r$ is the area of the rectangle and $S_D$ the area of the domain. As a
consequence of the independence of the random points, $k(r,\mathbf{X})$ follows
a binomial distribution. Thus, the probability term
$\mathds{P}\big[k(r,\mathbf{X})\geq k(r,\mathbf{x})\big]$ is given by
\begin{equation}
   \mathds{P}\Big[k(r,\mathbf{X}) \geq k(r,\mathbf{x})\Big] =
        \mathcal{B}\big(N,k(r,\mathbf{x}),p\big)
\end{equation}
where $\mathcal{B}(n,k,p)$ is the tail of the binomial distribution
\begin{equation}
   \mathcal{B}(n,k,p) = \sum_{j=k}^n \binom{n}{j}p^{j}(1-p)^{n-j}.
\end{equation}
The \emph{number of tests} $N_{tests}$ corresponds to the total number of
rectangles that could show an alignment, which in turn is related to the number
of pairs of points defining such rectangles. With a set of $N$ points this gives
$\frac{N\times (N-1)}{2}$ different pairs of points.

The set of rectangle widths to be tested must be specified \emph{a priori} as
well. In the \emph{a contrario} approach, a compromise must be found between the
number of tests and the precision of the gestalts that are being sought for. The
larger the number of tests, the lower the statistical relevance of
detections. However, if the set of tests is chosen wisely, gestalts fitting
accurately the tests will have a very low probability of occurrence under $H_0$
and will therefore be more significant.

At a digital image precision, the narrowest possible width for an alignment is 1
(taking the side of a pixel as length unit). The series of tested widths grows
geometrically until it achieves a maximal possible width, which can be set
\emph{a priori} as a function of the alignment length. Since the number of
tested widths depends on the length of the alignment, we cannot predict \emph{a
  priori} (before the dots have been drawn) how many tests will be
done. Fortunately the total number of widths can be estimated as the number of
widths tested in an average rectangle times the number of evaluated
rectangles. We call this quantity $W$. The impact of this approximation in the
detector results is insignificant \cite{DMM_book}. The total number of tested
rectangles is then:
\begin{equation}
   N_{tests} = \frac{ N(N-1)\times W }{2}.
\end{equation}
We will define now the fundamental quantity of the \emph{a contrario} framework,
the Number of False Alarms (NFA) associated with a rectangle $r$ and a set of
dots $\mathbf{x}$:
\begin{equation}\label{eq:basicNFA}
   \mathrm{NFA}(r,\mathbf{x})
      = N_{tests} \cdot\mathds{P}\Big[k(r,\mathbf{X})\geq k(r,\mathbf{x})\Big]
      = \frac{N(N-1)\times W}{2}\cdot\mathcal{B}\Big(N,k(r,\mathbf{x}),p\Big).
\end{equation}
This quantity corresponds, as said before
(Eq.~\ref{eq:expected_number_of_events}), to the expected number of rectangles
which have a sufficient number of \emph{points} to be as rare as $r$ under
$H_0$. When the NFA associated with a rectangle is large, this means that such
an event is to be expected under the \emph{a contrario} model, and therefore is
not relevant. On the other hand, when the NFA is small, the event is rare and
probably meaningful. A \emph{perceptual threshold} $\varepsilon$ must
nevertheless be fixed, and rectangles with
$\mathrm{NFA}(r,\mathbf{x})<\varepsilon$ will be called
\emph{$\varepsilon$-meaningful rectangles} \cite{es2003computational},
constituting the detection result of the algorithm.

\begin{theorem}[\cite{DMM_book}]
$$
   \mathds{E}\left[\sum_{R\in\mathcal{R}} \mathds{1}_{\mathrm{NFA}(R,\mathbf{X})<\varepsilon}
         \right] \leq \varepsilon
$$
where $\mathds{E}$ is the expectation operator, $\mathds{1}$ is the indicator
function, $\mathcal{R}$ is the set of rectangles considered, and $\mathbf{X}$ is
a random set of points on $H_0$.
\end{theorem}
The theorem states that the average number of \emph{$\varepsilon$-meaningful
  rectangles} under the \emph{a contrario} model $H_0$ is bounded by
$\varepsilon$. Thus, the number of detections in noise is controlled by
$\varepsilon$ and it can be made as small as desired. In other words, this shows
that our detector satisfies the non-accidentalness principle.

Following Desolneux, Moisan, and Morel \cite{dmm2000,DMM_book}, we shall set
$\varepsilon=1$ once and for all. This corresponds to accepting on average one
false detection per image in the \emph{a contrario} model, which is generally
reasonable. Also, the detection result is not sensitive to the value of
$\varepsilon$, see \cite{DMM_book}.

\begin{figure}[t]
\centering
\begin{tabular}{ccccc}
\fbox{\includegraphics[width=0.215\linewidth]{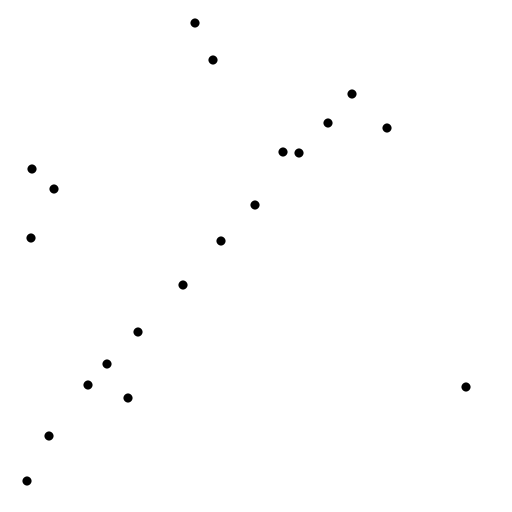}} &
\fbox{\includegraphics[width=0.215\linewidth]{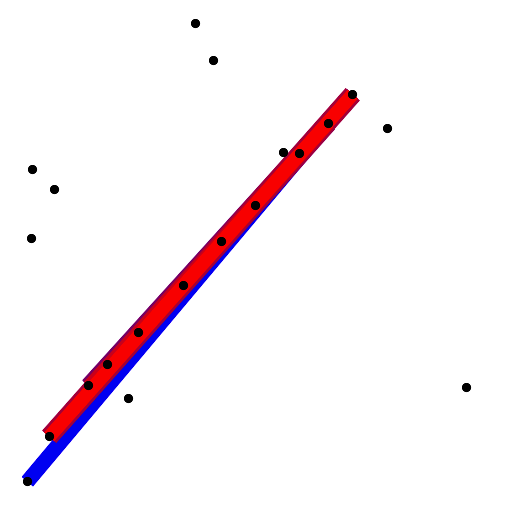}} &
\hspace{1mm} &
\fbox{\includegraphics[width=0.215\linewidth]{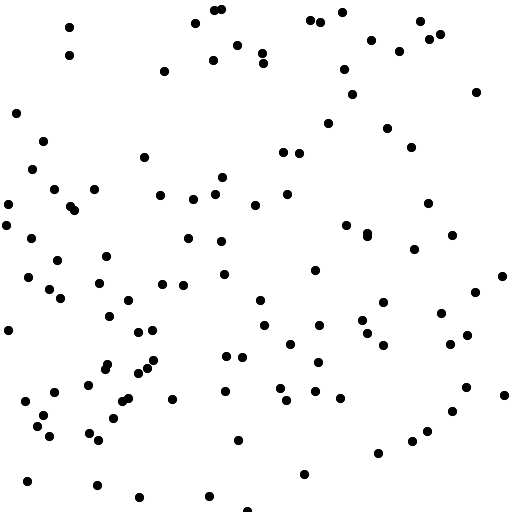}} &
\fbox{\includegraphics[width=0.215\linewidth]{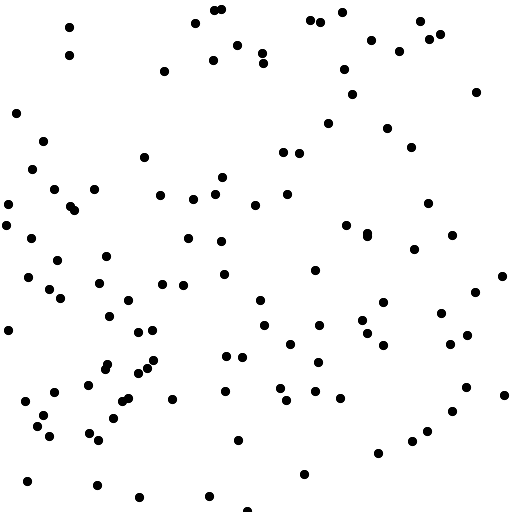}} \\
(a) & (b) & & (c) & (d) \\
\end{tabular}
\caption{Results from the basic dot alignment detector. \textbf{(a)} and
  \textbf{(c)} are the input data, and \textbf{(b)} and \textbf{(d)} are the
  corresponding results. Each detection is represented by a rectangle and its
  color indicates the NFA value. In \textbf{(b)} the algorithm correctly detects
  the obvious alignment. Notice that multiple and redundant rectangles were
  detected; this problem will be discussed in
  Sect.~\ref{sec:alignments_masking_principle}. The data set \textbf{(c)}
  contains the same set of points in \textbf{(a)} plus added noise dots, thus
  the aligned dots are still present. However, the algorithm handles correctly
  the masking by texture or noise and produces no detection.}
\label{fig:exp0}
\end{figure}

Figure~\ref{fig:exp0} shows the results of the basic algorithm in two simple
cases. The results are as expected: the visible alignment in the first example
is detected, while no detection is produced in the second. Actually, the dots in
the first example are also present in the second one, but the addition of random
dots masks the alignment, in accordance with human perception. Note that the
first example produces many redundant detections; this problem will be handled
in Sect.~\ref{sec:alignments_masking_principle}.

\subsection{A Refined Dot Alignment Detector}\label{sec:refined-align}

Naturally, the simple model for dot alignment detection presented in the last
section does not take into account many situations that can arise and
significantly affect the perception of alignments. For example: what happens if
there are point clusters inside the alignment? What if the background image has
a non uniform density? Should not the algorithm prefer alignments where the
points are equally spaced? These questions, among others, arise when subjects
play the gestaltic game and try to fool the algorithm with new drawings. There
are two ways to fool the algorithm: One is by drawing a particular context that
prevents the algorithms from detecting a conspicuous alignment. Inversely, the
other sort of counterexample is a drawing inducing detections that remain
invisible to the human eye. As more counterexamples are found, more
sophisticated versions of the algorithm must be developed, and each new version
will become harder to falsify than the previous one.

Using this methodology, we produced several refined versions of the basic
algorithm. Here we will present the principal counterexamples that were found,
and then describe the last version of the algorithm which takes all of them into
account. This algorithm is therefore harder to fool. Ideally, the game should
end when the \emph{Turing test} \cite{pinar2000turing} is satisfied, namely when
a human observer will be unable to distinguish between the detections produced
by a machine and by a human.

\begin{figure}[t]
\begin{center}
\begin{tabular}{ccc}
\fbox{\includegraphics[width=0.3\textwidth]{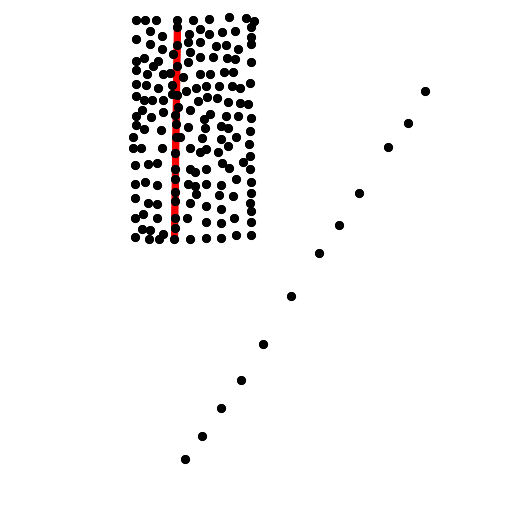}} &
\fbox{\includegraphics[width=0.3\textwidth]{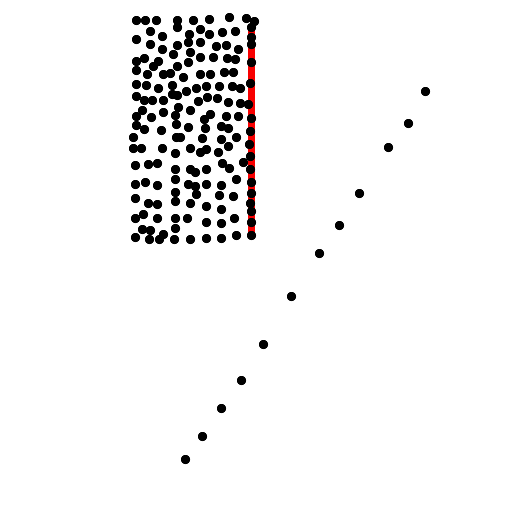}} &
\fbox{\includegraphics[width=0.3\textwidth]{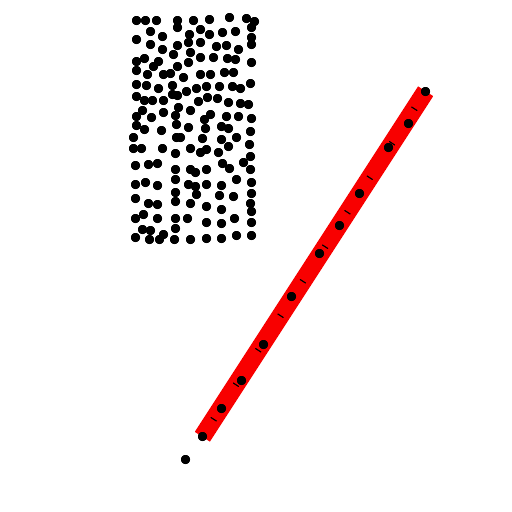}}\\
(a) & (b) & (c) \\
\end{tabular}
\end{center}
\caption{Local vs. global density estimation. In each example, only the most
  meaningful detected alignment (the one with the lowest NFA) is shown for each
  algorithm. The algorithms (a), (b), (c) use a background model with growing
  complexity to avoid wrong detections. \textbf{(a)} global density estimation:
  the detected segment is not the most meaningful for our perception, but has
  nevertheless a high dot density compared to the average image density used as
  background model. \textbf{(b)} here a local density estimation gives the
  background model, but the local density is lower on the border of the big dot
  rectangle, hence the detection. \textbf{(c)} this last problem is avoided by
  computing a local density estimation taking the maximum density on both sides
  of the alignment.}
\label{fig:local_vs_global}
\end{figure}

First, we noticed a deficiency in the detector when zones in the image have
higher dots density. This problem arises naturally from the wrong \emph{a
  contrario} assumption that the whole image has the same density of
points. When it is not the case, the global density estimation can be misleading
and produces poor detection results, as illustrated in
Fig.~\ref{fig:local_vs_global}~(a). The solution for this is to compute a
\emph{local density estimation} with respect to the evaluated rectangle. The
algorithm uses a local window with size proportional to the width of the
evaluated alignment.

However, this local density estimation can introduce new problems such as a
``border effect'', as shown in Fig.~\ref{fig:local_vs_global}~(b). Indeed, the
density estimation is lower on the border of the dot rectangle than inside it,
because outside the rectangle there are no dots. Thus, the algorithm detects on
the border a non-accidental, meaningful excess with respect to the local
density.

\begin{figure}[t]
\centering
\fbox{\includegraphics[width=.4\linewidth]{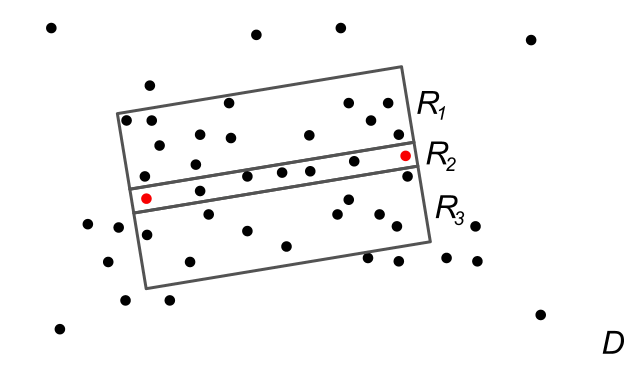}}
\caption{In the refined version of the algorithm, the density of points is
  measured to each side of the evaluated rectangle. The maximum of the densities
  in $R_1$ and $R_3$ is taken and this value is used as an estimation of the dot
  density in both $R_1$ and $R_3$.}
\label{fig:max_density_schema}
\end{figure}

In order to avoid this effect, the version of the algorithm used in
Fig.~\ref{fig:local_vs_global}~(c) measures for the background model the
\emph{maximum of the densities} measured on both sides of the alignment. In
short, to be detected, an alignment must show a higher dot density than in both
regions immediately on its left and right. This local alignment detector is
therefore similar to classic second order Gabor filters where an elongated
excitatory region is surrounded by two inhibitory regions. The local points
estimation is calculated in the following way, see
Fig.~\ref{fig:max_density_schema}. The local window is divided in three
parts. $R_1$ is the rectangle formed by the area of the local window on the left
of the alignment. $R_3$ is the area of the local window on the right of the
alignment, and $R_2$ is the rectangle which forms the candidate alignment. Note
that the length of the local window is the same as the alignment and that we can
consider any arbitrary orientation for it. Next, the algorithm counts the
numbers of dots $M_1$, $M_2$, and $M_3$ in $R_1$, $R_2$ and $R_3$
respectively. Finally the \emph{a contrario} model assumes that the number of
dots in the local window $R_1\cup R_2\cup R_3$ is
\begin{equation}\label{eq:max_density}
   n(r,\mathbf{x}) = \max(M_1,M_3) \times 2 + M_2,
\end{equation}
and that these dots are randomly distributed.

\begin{figure}[b]
\begin{center}
\begin{tabular}{ccc}
\fbox{\includegraphics[width=0.3\textwidth]%
                      {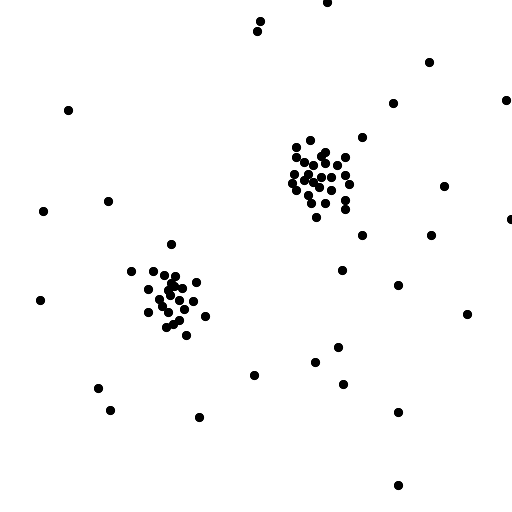}} &
\fbox{\includegraphics[width=0.3\textwidth]{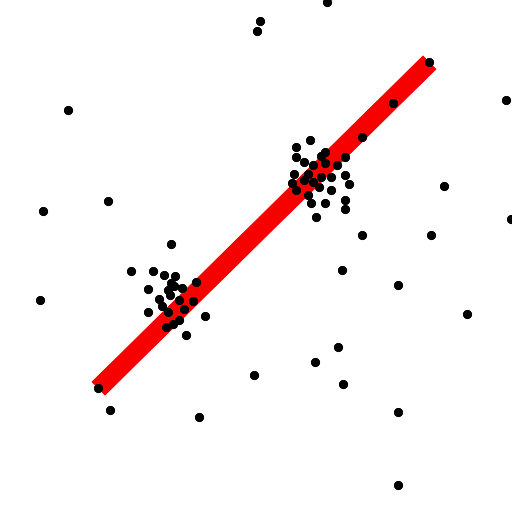}} &
\fbox{\includegraphics[width=0.3\textwidth]{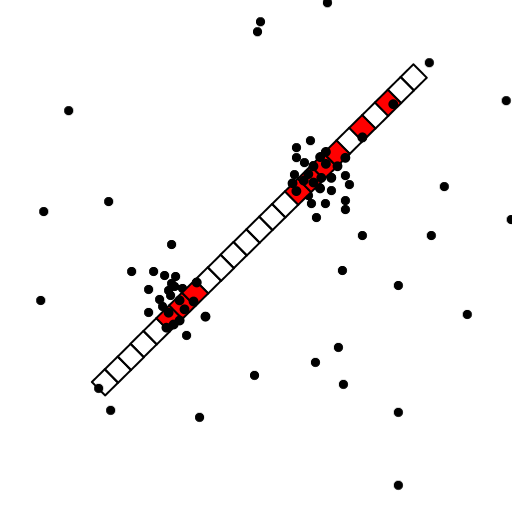}} \\
(a) & (b) & (c) \\
\end{tabular}
\end{center}
\caption{Counting occupied boxes to avoid false detections from the presence of
  clusters. The dot pattern shown in image \textbf{(a)} presents two dot
  clusters but no alignment. However, the basic algorithm finds a thin rectangle
  with a high dot density, hence a false detection, as shown in
  \textbf{(b)}. Dividing the rectangle into boxes and counting the occupied
  ones, avoids this misleading cluster effect, as seen in \textbf{(c)}, where
  the occupied boxes are marked in red and no alignment is actually detected.}
\label{fig:small_boxes}
\end{figure}

There is still an objection to this new algorithm, obtained in the gestaltic
game by introducing small dot clusters, as shown in
Fig.~\ref{fig:small_boxes}~(a). The detected alignment in
Fig.~\ref{fig:small_boxes}~(b) seems clearly wrong. There is indeed a meaningful
dot density excess inside the red rectangle, but this excess is caused by the
clusters, not by what could be termed an alignment. While the algorithm counted
every point, the human perception seems to group the small clusters into a
single entity, and count them only once. Also, as suggested in other studies
\cite{PreissThesis,Tripathy99,Uttal73}, the density is not the only property
that makes an alignment perceptually meaningful; another characteristic to
consider is the uniform spacing of the dots in it, which the gestaltists call
the principle of \emph{constant spacing}. These objections have led to a still
more sophisticated version of the alignment detector. In order to take into
account both issues (avoiding small clusters and favoring regular spacing) a
more advanced version of the alignment detector was designed which divides each
candidate rectangle into equal \emph{boxes}. The algorithm counts the number of
boxes that are occupied by at least one point, instead of counting the total
number of points. In this way, the minimal NFA is attained when the dots are
perfectly distributed along the alignment. In addition, a concentrated cluster
in the alignment has no more influence on the alignment detection than a single
dot in the same position.

The NFA calculation for this refined version of the algorithm is slightly
different than for the basic one. The event for which we are estimating an
expected number of occurrences in a background model is defined as
follows. Given two points and a number of boxes $c$, the question is: What is
the probability that the number of occupied boxes among the $c$ is larger than
the expected number under the \emph{a contrario} model? Let us start by
computing the probability of one dot falling in one of the boxes:
\begin{equation}
   p_0 = \frac{S_B}{S_L},
\end{equation}
where $S_B$ and $S_L$ are the areas of the boxes and the local window
respectively. Then, the probability of having one box occupied by at least one
of the $n(r,\mathbf{x})$ dots (Eq.~\ref{eq:max_density}) is:
\begin{equation}\label{eq:p1_small_boxes}
   p_1 = \mathcal{B}\big(n(r,\mathbf{x}),1,p_0\big).
\end{equation}
We call \emph{occupied} boxes the ones that have at least one dot inside, and we
will denote by $b(r,c,\mathbf{x})$ the observed number of occupied boxes in the
rectangle $r$ divided into $c$ boxes. Finally, the probability of having at
least $b(r,c,\mathbf{x})$ of the $c$ boxes occupied is
\begin{equation}\label{eq:p2_small_boxes}
   \mathcal{B}\big(c,b(r,c,\mathbf{x}),p_1\big).
\end{equation}
A set $\mathcal{C}$ of different values are tried for the number of boxes $c$
into which the rectangle is divided. Thus, the number of tests needs to be
multiplied by its cardinal $|\mathcal{C}|$. In practice we set
$|\mathcal{C}|=\sqrt{N}$ and that leads to
\begin{equation}
   N_{tests} = \frac{ N(N-1)\times W\times |\mathcal{C}| }{2}
           = \frac{ N(N-1)\times W\times \sqrt{N} }{2}.
\end{equation}
The NFA of the new event definition is then:
\begin{equation}\label{eq:small_boxes}
   \mathrm{NFA}(r,\mathbf{x}) = \frac{ N(N-1)\times W\times \sqrt{N} }{2} \cdot
                \min_{c\in\mathcal{C}} \mathcal{B}\big(c,b(r,c,\mathbf{x}),p_1\big).
\end{equation}

Figs.~\ref{fig:local_vs_global}~(c) and~\ref{fig:small_boxes}~(c) show two
examples of the resulting algorithm, and we will show some more after discussing
the masking problem.

\subsection{Masking}\label{sec:alignments_masking_principle}

\begin{figure}[t]
\begin{center}
\begin{tabular}{ccc}
\fbox{\includegraphics[width=.3\textwidth]{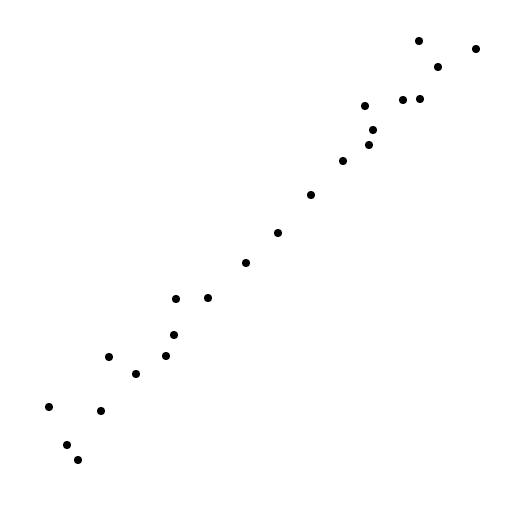}}&
\fbox{\includegraphics[width=.3\textwidth]{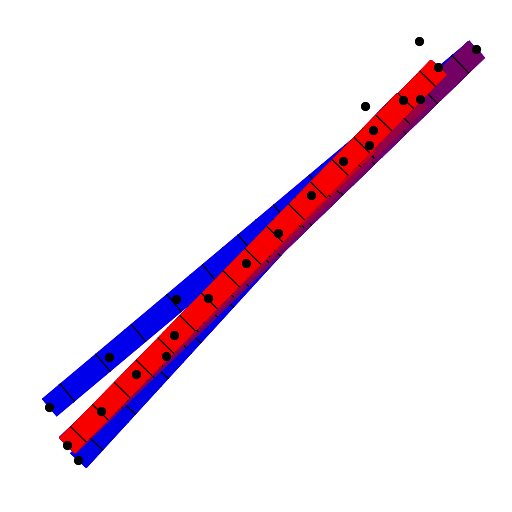}}&
\fbox{\includegraphics[width=.3\textwidth]{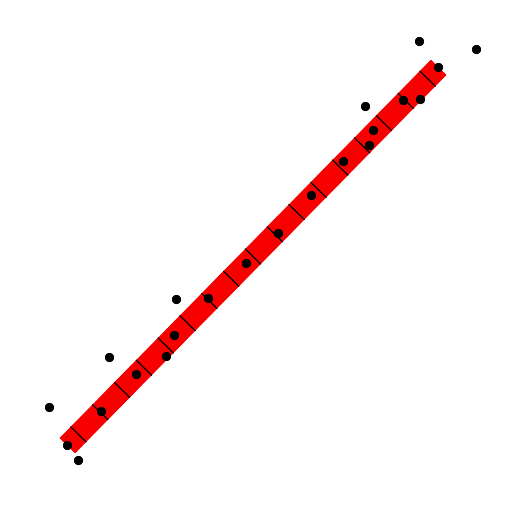}}\\
\end{tabular}
\end{center}
\caption{Redundant detections. \textbf{Left:} dot pattern. \textbf{Center:} all
  significant alignments found by the refined dot alignment detector described
  in Sect.~\ref{sec:refined-align}. The color represents the relative NFA value,
  where red is the most significant (smallest NFA value) and blue the least
  (highest NFA value). \textbf{Right:} Result of the masking process.}
\label{fig:nonmaxsup}
\end{figure}

As was observed in Fig.~\ref{fig:exp0}, all the described alignment detectors
may produce redundant detections. The reason is that a relevant gestalt is
generally formed by numerous elements and many subgroups also form relevant
gestalts in the sense of the non-accidentalness principle. Every pair of dots
defines a rectangle to be tested. Clearly, in a conspicuous alignment there will
be many such rectangles that partially cover the main alignment and are
therefore also meaningful. This redundancy phenomenon can involve dots that
belong to the real alignment as well as background dots near the alignment, that
can contribute to a rectangle containing a large number of dots, as illustrated
in Fig.~\ref{fig:nonmaxsup}. However, in such cases humans perceive only one
gestalt. Indeed, one could expect that there is only one causal reason leading
to redundant detections and it makes sense to select the best rectangle to
represent it.

\begin{figure}[b]
\begin{center}
\includegraphics[width=.45\textwidth]{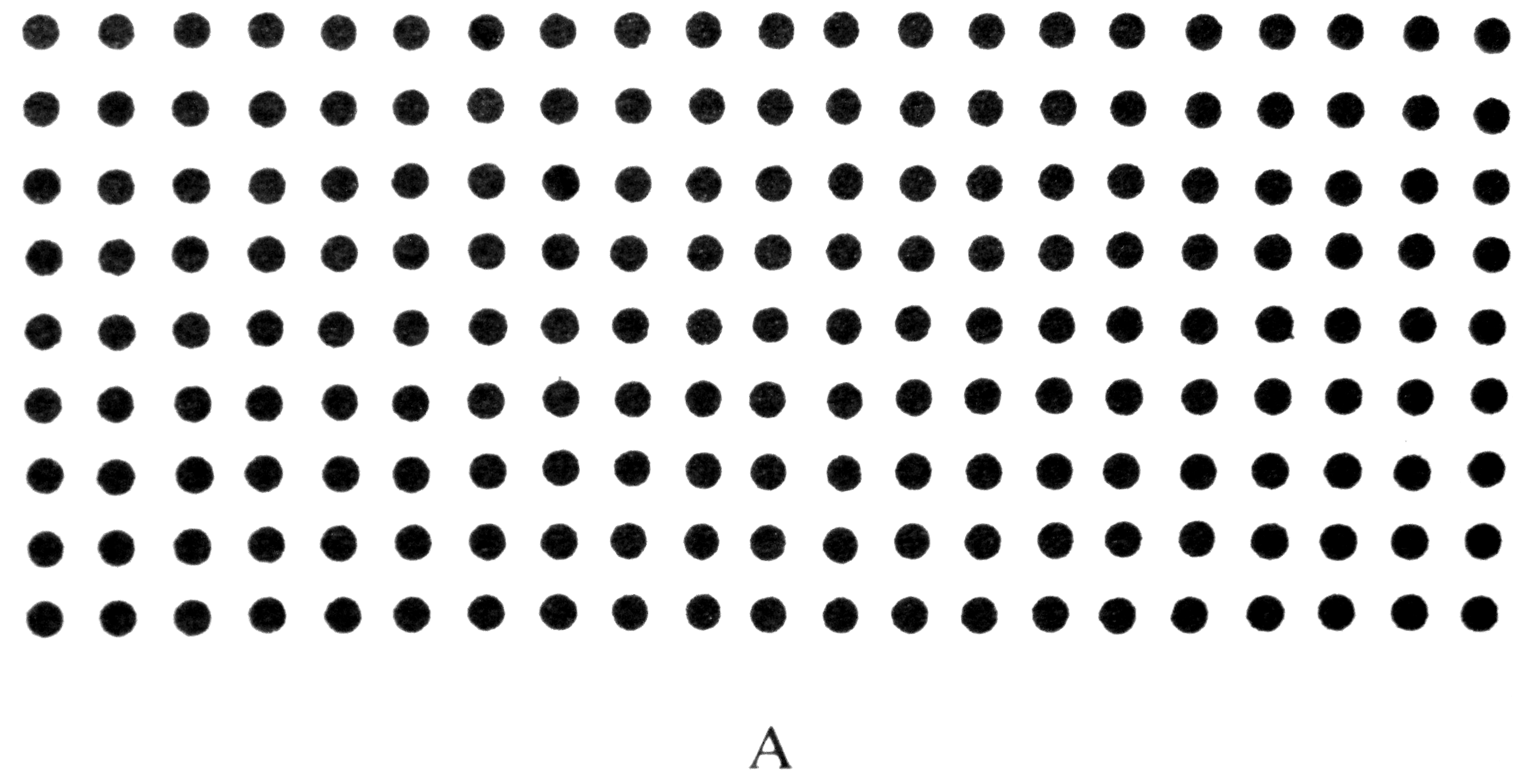}
\hspace{1cm}
\includegraphics[width=.35\textwidth]{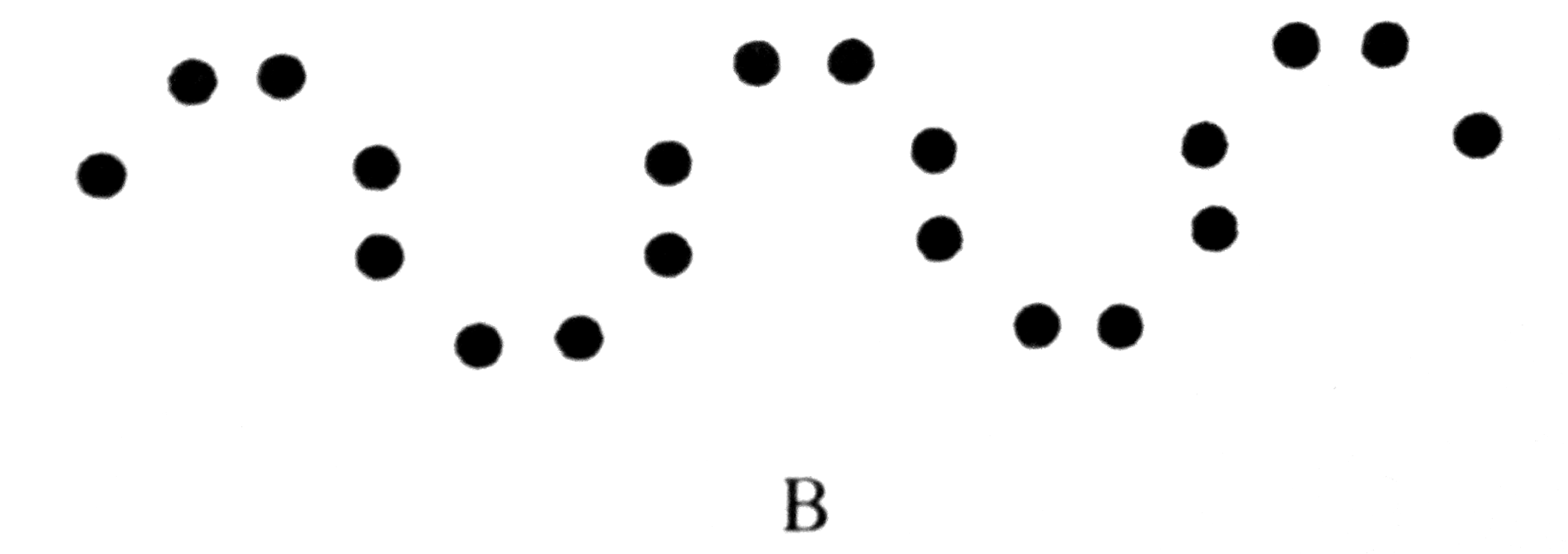}
\end{center}
\caption{A masking example by Kanizsa \cite[p.155]{kanizsa:vedere}: The
  ``curve'' in B is also present in the grid of dots A; nevertheless, it is not
  visible as it is masked by the global matrix configuration.}
\label{fig:masking}
\end{figure}

A similar phenomenon is described in the Gestalt literature
\cite{kanizsa:vedere}. Most scenes contain other possible interpretations that
are masked by the global interpretation. A simple example is shown in
Fig.~\ref{fig:masking} where subsets of the grid of dots form a huge quantity of
gestalts, but are invisible because they are masked by the rectangular matrix of
dots. This fact is, after Vicario, called Kanizsa's paradox \cite{DMM_book}.

A simple model for this masking process was proposed by
Desolneux~et~al. \cite{DMM_book} under the name of ``exclusion principle''. The
main idea is that each \emph{basic element} (for example the dots) cannot
contribute to more than one perceived group or gestalt. The process is as
follows: The most meaningful observed gestalt (the one with smallest NFA) is
kept as a valid detection. Then, all the basic elements (the dots in our case)
that were part of that validated group are assigned to it and the remaining
candidate gestalts cannot use them anymore. The NFA of the remaining candidates
is re-computed without counting the excluded elements. In that way redundant
gestalts lose most of their supporting elements and are no longer meaningful. On
the other hand, a candidate that corresponds to a different gestalt keeps most
or all of its supporting basic elements and remains meaningful. The most
meaningful candidate among the remaining ones is then validated and the process
is iterated until there are no more meaningful candidates.

\begin{figure}[b]
\begin{center}
\fbox{\includegraphics[width=.3\textwidth]%
                      {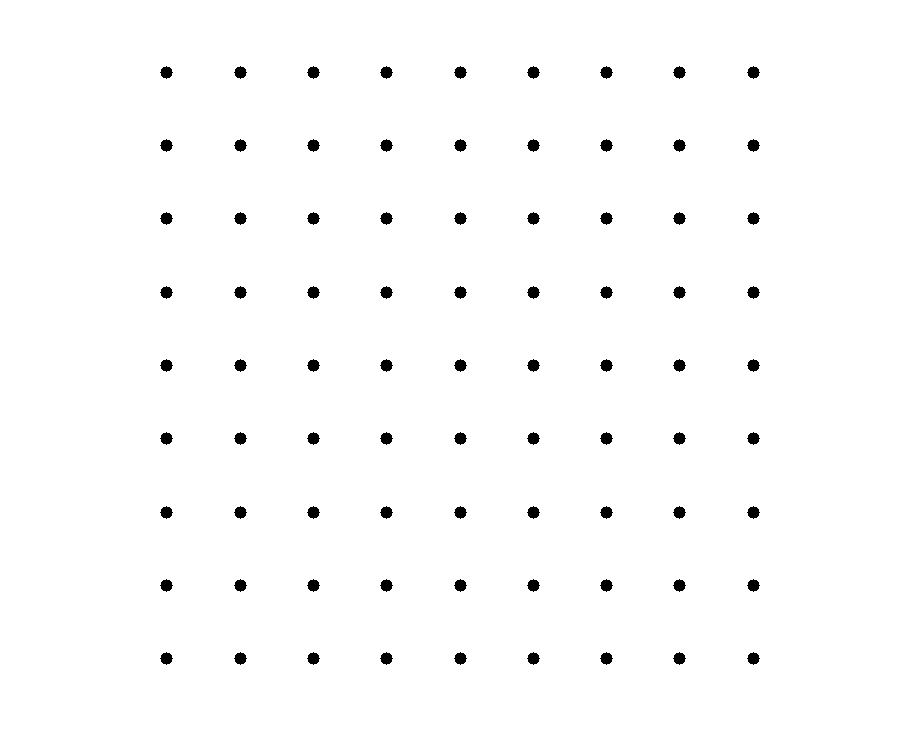}}
\fbox{\includegraphics[width=.3\textwidth]{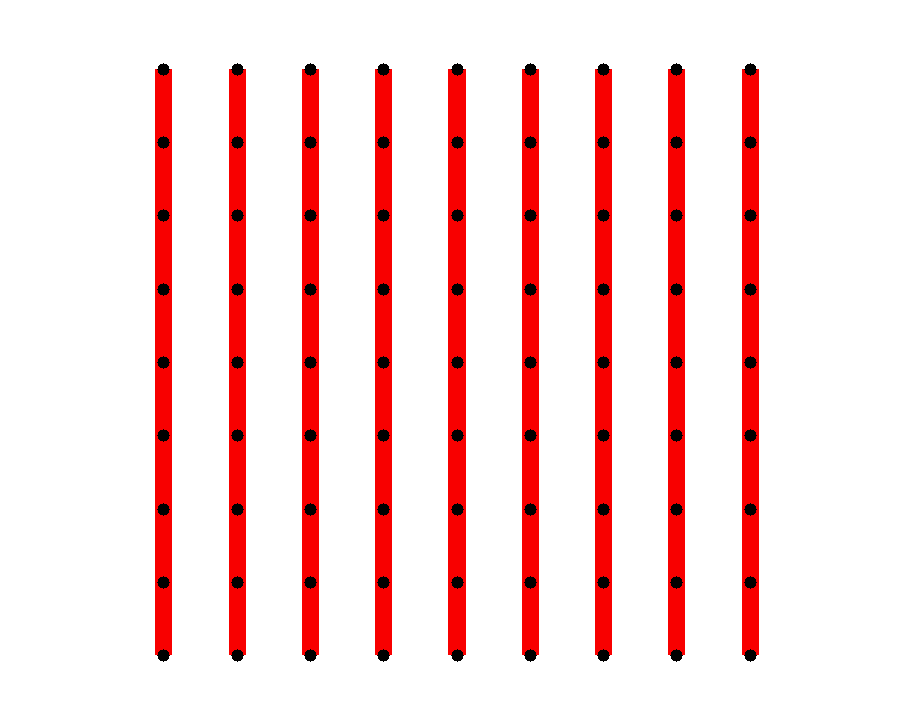}}
\fbox{\includegraphics[width=.3\textwidth]{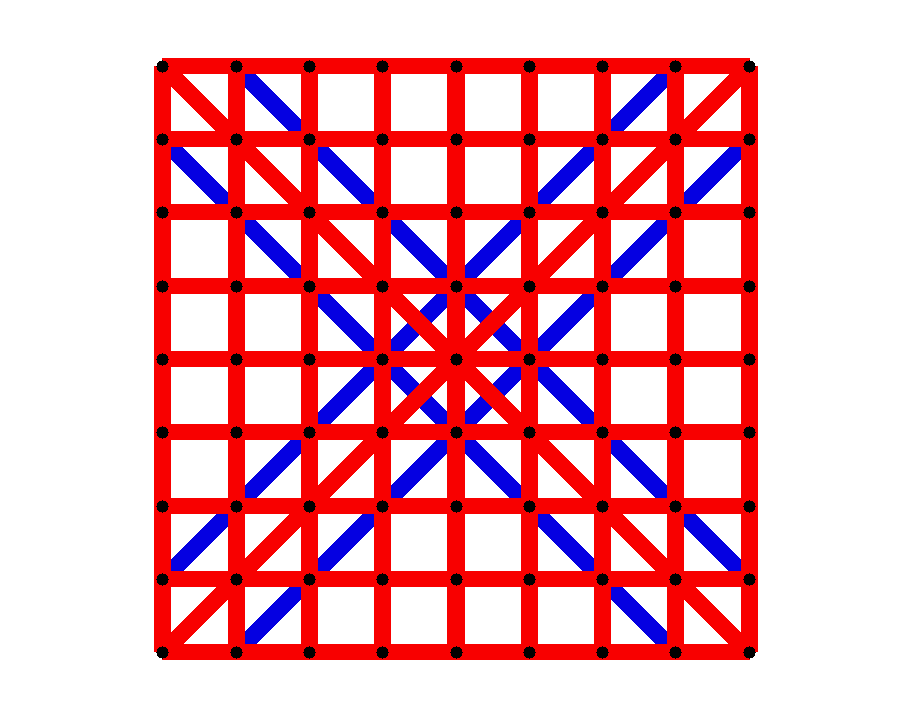}}
\end{center}
\caption{Examples of two alternative formulations of the masking
  process. \textbf{Left:} Set of dots. \textbf{Center:} The Exclusion Principle
  as defined in \cite{DMM_book}, a validated gestalt prevents others from using
  its dots. The vertical alignments (that were evaluated first) mask almost all
  the horizontal ones. \textbf{Right:} The Masking Principle, described in the
  text, which solves the ambiguities without forbidding basic elements to
  participate of two different gestalts. In this example, no individual
  alignment can mask an individual one in another direction. Thus we get all
  oblique, horizontal and vertical meaningful alignments.}
\label{fig:1vs1}
\end{figure}

This formulation of the masking process often leads to good results, removing
redundant detections while keeping the good ones. However, the gestaltic game
showed that it may also lead to unsatisfactory results as illustrated in
Fig.~\ref{fig:1vs1}. The problem arises when various gestalts have many elements
in common. As one gestalt is evaluated after the other, it may happen that all
of its elements have been removed, even if the gestalt is in fact not redundant
with any of the other ones. In the example of Fig.~\ref{fig:1vs1}, individual
horizontal and vertical alignments are not redundant, but if all the vertical
ones have been detected first, the remaining horizontal ones will be
(incorrectly) masked. This example shows a fundamental flaw of the exclusion
principle: it is not sound to impose that a basic element belongs to a single
perceptually valid gestalt. There must be a global explanation of the
organization of the basic elements in visible gestalts which is at the same time
coherent with each individual gestalt (eliminating local redundancy) and with
the general explanation of the scene in such a way that some basic elements can
participate of several gestalts without contradiction. The solution seems to be
in a sort of relaxation of the exclusion principle. The following definitions
sketch a possible solution.

\begin{definition}[Building Elements]\label{basic-element}
We call \emph{building element} any atomic image component that can be a
constituent element of several gestalts. Valid examples of building elements are
dots, segments, or even gestalts themselves, that can be recursively grouped in
clusters or alignments. From that point of view any gestalt can be used as a
building element for higher level gestalts.
\end{definition}

\begin{definition}[Masking Principle]\label{masking-principle}
A meaningful gestalt $B$ will be said ``masked by a gestalt $A$'' if $B$ is no
longer meaningful when evaluated without counting its building elements
belonging to $A$. In such a situation, the gestalt is not retained as detected.
\end{definition}

\begin{figure}[b]
\begin{center}
\fbox{\includegraphics[width=.22\textwidth]%
                      {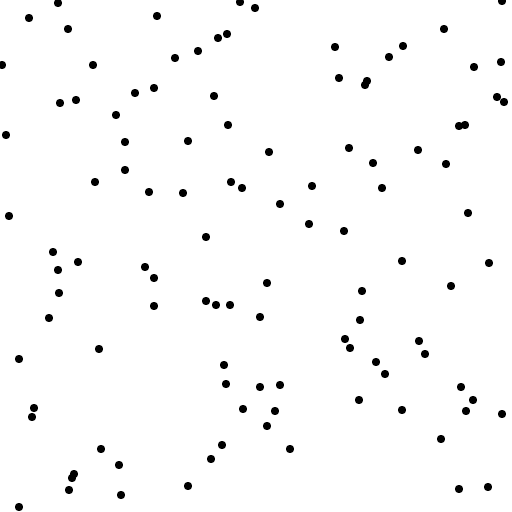}}
\fbox{\includegraphics[width=.22\textwidth]%
                      {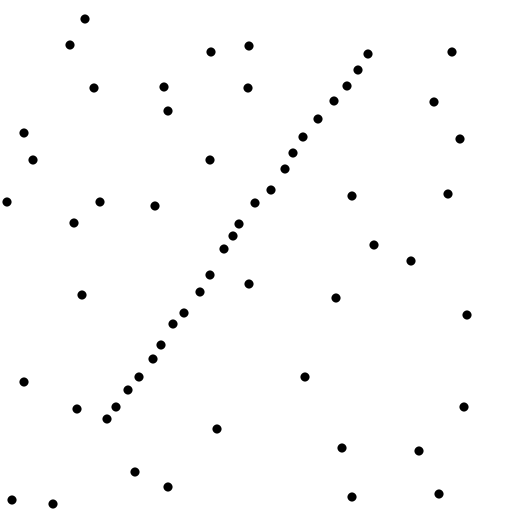}}
\fbox{\includegraphics[width=.22\textwidth]%
                      {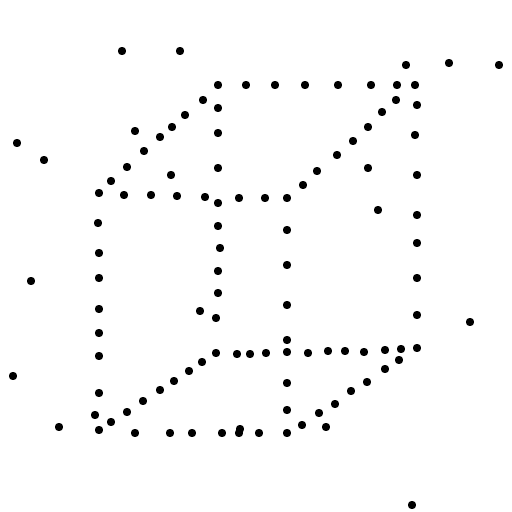}}
\fbox{\includegraphics[width=.22\textwidth]%
                      {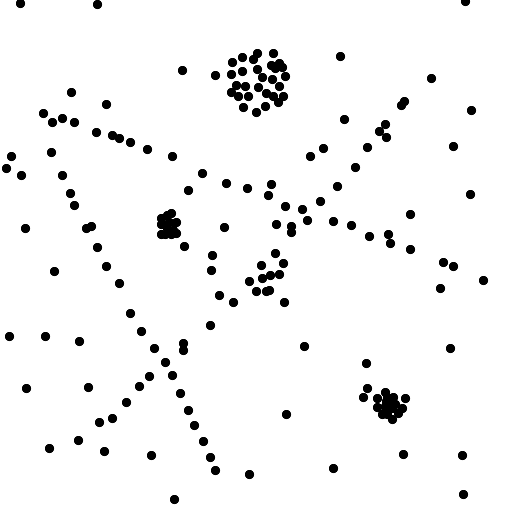}}
\fbox{\includegraphics[width=.22\textwidth]{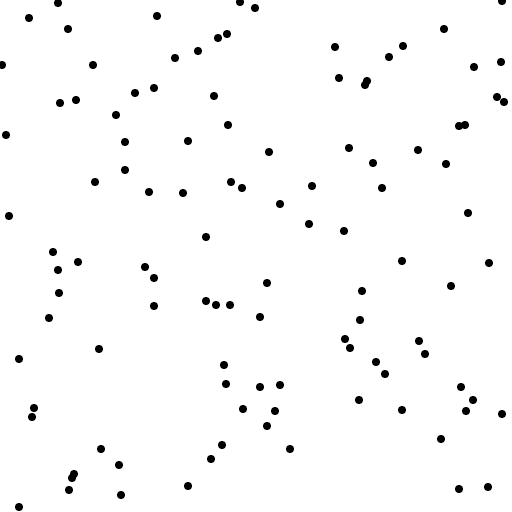}}
\fbox{\includegraphics[width=.22\textwidth]{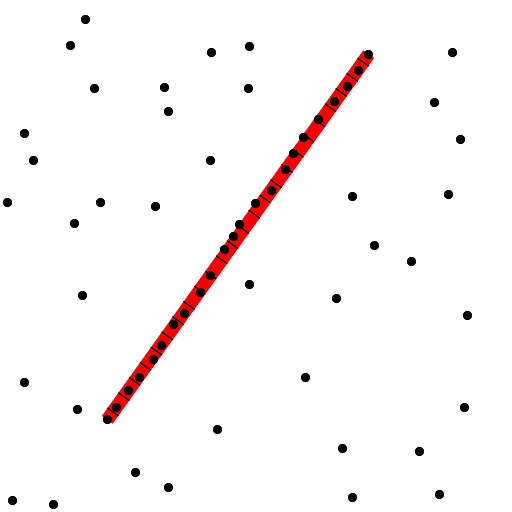}}
\fbox{\includegraphics[width=.22\textwidth]{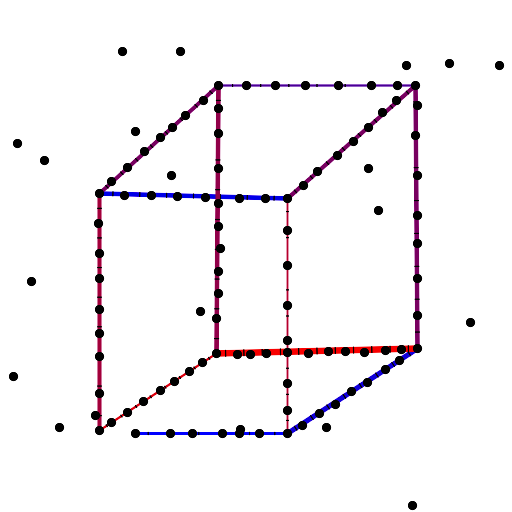}}
\fbox{\includegraphics[width=.22\textwidth]{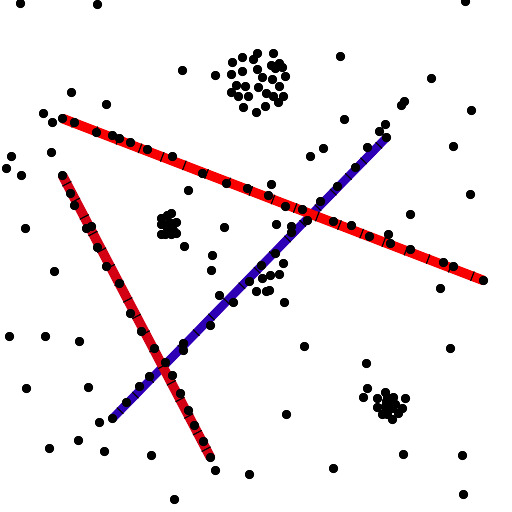}}
\end{center}
\caption{Results of the final dot alignment detector, using the refined method
  described in Sect.~\ref{sec:refined-align} in conjunction with the Masking
  Principle (Def.~\ref{masking-principle}). The top row is the input data; the
  bottom row shows the results.}
\label{fig:final_examples}
\end{figure}

In short, a meaningful gestalt will be detected \emph{if} it is not masked by
any other detected gestalt. The difference is that here a gestalt can only be
masked by another \emph{individual} gestalt and not by the union of several
gestalts as is possible with the exclusion principle. Thus this masking
principle is analogous to a Nash equilibrium, in the sense that every gestalt
remains meaningful when separately subtracting from it the building blocks of
any other gestalt. A procedural way to attain this result is to validate
gestalts one by one, starting by the one with smallest NFA; before accepting a
new gestalt, it is checked that it is not masked by any one of the previously
detected gestalts. The masking principle applies easily to point alignments.

Fig.~\ref{fig:final_examples} shows some dot alignment detection results when
combining the method of the previous section and the masking principle. The
results obtained in these examples are as expected.

\subsection{Online Gestaltic Game}

The gestaltic game allowed us to discover examples of dot arrangements that the
current algorithm is not able to handle correctly. The hardest ones we
encountered to date belong to the ``masking by structure'' kind, as those
presented in the right hand part of Fig.~\ref{fig:masking_examples}. Surely
there are more cases than those discovered so far. To facilitate the search we
created an online interface where everyone can easily play the gestaltic game
inventing new
counterexamples.\footnote{\url{http://dev.ipol.im/~jlezama/dot_alignments}}

Being interactive, the online gestaltic game is designed to be eventually
published in the IPOL journal. It allows users to draw their own dot patterns
and to see the output of the detection algorithm. Alternatively, the user can
upload a set of dots, or modify an existing one by adding or removing individual
dots or adding random dots. All the experiments are stored and accessible in the
``archive'' part of the site and may help improve the theory.

Current work is focused on the conflict between different gestalts with the
objective of handling the masking by structure problem.

\section{Detection Theory versus Psychophysics}\label{sec:psychophysics}

In this second part we leave the question of a quantitative gestaltism and go
back to more classic psychophysics. The question is whether a quantitative
framework like the \emph{a contrario} detection theory can also become a useful
addition for human contour perception psychophysical experiments.

\begin{figure}[t]
\centering
\includegraphics[width=0.3\textwidth]{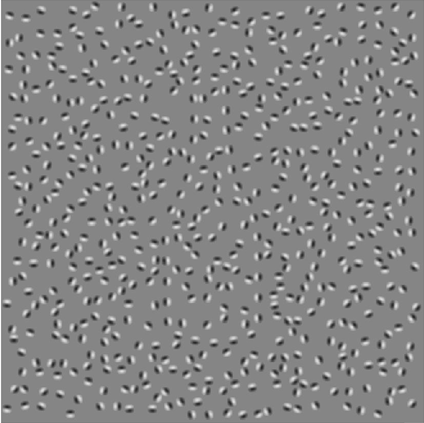}
\includegraphics[width=0.3\textwidth]{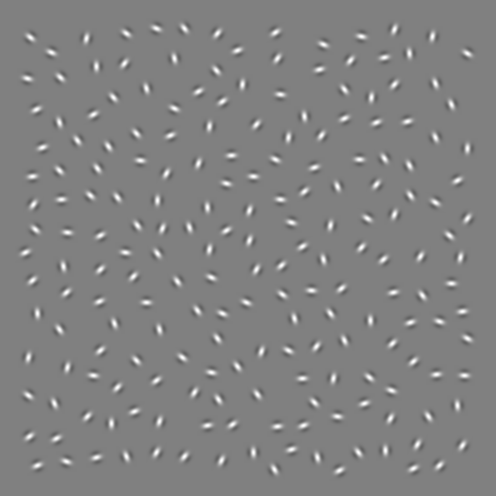}
\caption{\textbf{Left:} An image extracted from
  Nyg{\aa}rd~et~al. \cite{Nygard09orientation_jitter_motion}. \textbf{Right:}
  Example of an alignment detection experiment to be developed here.}
\label{fig:wagemans}
\end{figure}

Arrays of Gabor patches have become a classic tool for the study of the
influence of good continuation in perceptual
grouping~\cite{Field93ContourIntegration,
  Nygard09orientation_jitter_motion}. Gabor functions ensure a control on the
stimuli spectral complexity and on the spatial scale of the contours. They give
a flexible and easy way for building a great variety of stimuli. It has been
verified that the more aligned the Gabor patches are to the contour they lie on,
the easier their perceptual grouping into a shape's
outline~\cite{Field93ContourIntegration,
  Nygard09orientation_jitter_motion}. Fig.~\ref{fig:wagemans} (left), shows an
easy example where most subjects recognize a bottle. But the more freedom is
left to the Gabor orientation, the harder it is to distinguish such contours
from the background. For the influence of other perturbations of the contour
such as its motion or its curvature on the object's identifiability, we refer to
a recent study \cite{Nygard09orientation_jitter_motion}.

Can we hope for a quantitative interpretation to this experimental framework,
namely a function of the stimuli parameters that would predict and explain the
evolution of the detection performance? Probabilistic approaches (mainly
Bayesian) exist for contour modeling from a perceptual point of view
\cite{Attneave54InformationalAspects, Feldman05InfoAlongContours}, and have
sometimes been compared experimentally to human visual perception
\cite{Feldman01BayesianContour}; but none of these approaches proposed to
compute \emph{a priori} detection thresholds as functions of the stimuli
parameters.

The influence of experimental factors such as the length of the alignment, the
density of the patches, and the angular accuracy on human detection is a classic
subject of psychophysical inquiry. But the question of whether human performance
can be measured with only one adequate quantitative function of the parameters
is still open. We shall explore here if the NFA furnished by the \emph{a
  contrario} theory can play this role. Indeed, the NFA retains the remarkable
property of being a scalar function of the three psychophysical parameters
generally used in this kind of detection experiment. In classic experimental
settings, these parameters are varied separately and independently, and no
synthetic conclusion can be drawn; only separate conclusions on the influence of
each parameter can be reached. If a function like the NFA could play the role of
generic \emph{detectability} parameter, the experimental parameters could for
example be made to vary simultaneously in the very same experiment. In short, if
the hypothesis of a single underlying detection parameter is validated, this
would simplify the experimental setups and entail a new sort of quantitative
analysis of the results, two stimuli being \emph{a priori} considered as
equivalent in difficulty if their NFA are similar.

The underlying hypothesis, that the reaction of the subjects to varying stimuli
might be predicted as a single scalar function of the stimulus' parameters, is
equivalent to the classic hypothesis of a ``single mechanism'' for contour
detection. More precisely, we shall explore if this single mechanism might obey
the non-accidentalness principle (the NFA being its probabilistic quantitative
expression).

To keep the line of the previous section, this study will again focus on the
same simple gestalt: straight contours, that is to say alignments of Gabor
elements, as illustrated in Fig.~\ref{fig:wagemans} (right). The remainder of
this section describes the patterns used, the \emph{a contrario} method, the
experiment performed on humans, and the result of the comparison.

\subsection{The Patterns}\label{sec:patterns}

\begin{figure}[b]
\centering
\begin{tabular}{ccc}
\includegraphics[width=0.3\textwidth]{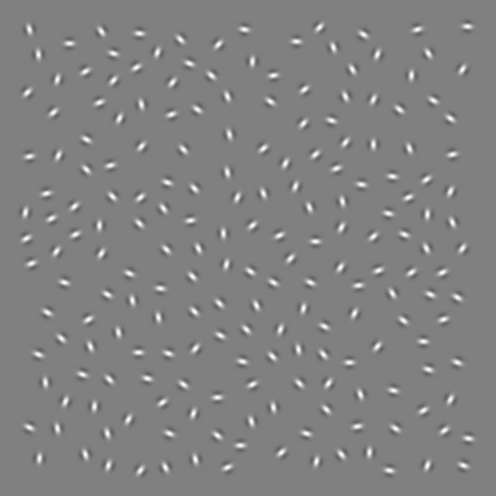} &
\includegraphics[width=0.3\textwidth]{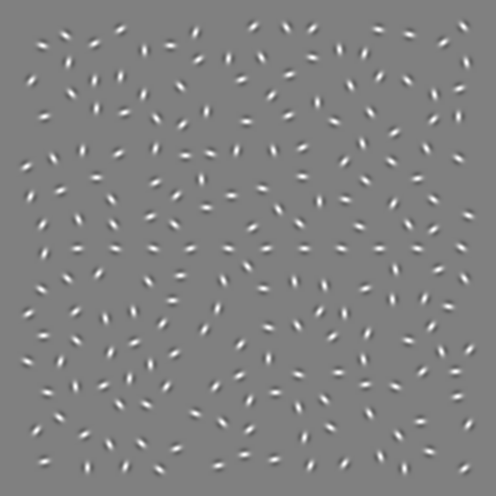} &
\includegraphics[width=0.3\textwidth]{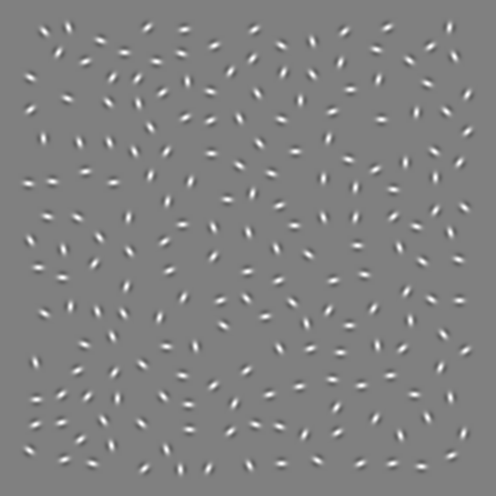} \\
(a) & (b) & (c) \\
\end{tabular}
\caption{Three examples of stimuli used in our experiments. \textbf{(a)} A
  jitter-free alignment with 10 elements. \textbf{(b)} A weakly jittered
  alignment with 10 elements. \textbf{(c)} A stimulus with no alignment,
  containing only elements with random orientations.}
\label{fig:EcvpStimuliExamples}
\end{figure}

Figure~\ref{fig:EcvpStimuliExamples} shows three examples of the stimuli used in
our experiments. All of them consist of symmetric Gabor elements with varying
positions and orientations placed over a gray background. There are two kinds of
stimuli: positive stimuli and negative stimuli. Negative stimuli contain
elements with random orientations sampled in $[0, \pi)$, e.g.
Fig.~\ref{fig:EcvpStimuliExamples}~(c). Positive stimuli, see
Fig.~\ref{fig:EcvpStimuliExamples}~(a) and~(b), contain a majority of random
elements like in negative stimuli but also a small set of \emph{foreground}
elements. The latter lie on a straight line and are uniformly spaced; their
orientations are randomly and uniformly sampled from an interval centered on the
alignment direction. The size of this interval gives a measure of the
\emph{angular jitter} and will be noted by $J$. When the jitter is zero, the
foreground elements have the exact same orientation as their supporting
line. Inversely, a jitter of $\pi$ leads to completely isotropic elements.

\begin{figure}[t]
\centering
\includegraphics[width=0.3\textwidth]{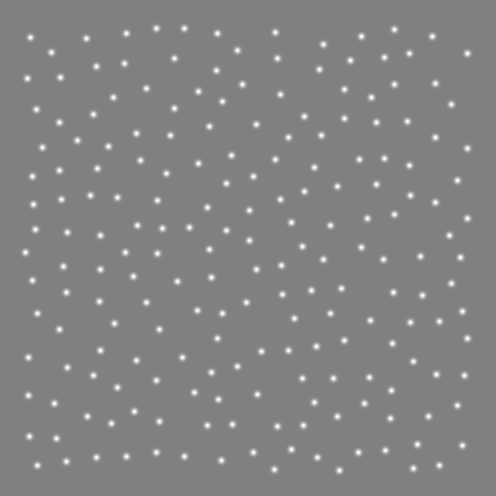}
\caption{Influence of stimuli position. Each dot represents the position where a
  Gabor element will be placed. The figure includes a perfectly regular and
  aligned set of dots, surrounded by random placed elements, all generated by
  the GERT package. It is very difficult to find the alignment, which shows that
  the position of the elements by itself conveys few cues about the presence of
  the alignment. (For a comparison, see the same stimulus with Gabor elements,
  Fig.~\ref{fig:wagemans} (right), where the alignment is easily spotted.)}
\label{fig:gabor-position}
\end{figure}

The experiment is designed to study how \emph{angular} jitter affects
visibility. Yet, a natural question arises about the contribution to the
detection of the accuracy of the alignment and of the regular spacing of the
aligned elements. All the stimuli presented in this section were generated with
the software GERT (v1.1) that includes special algorithms for the generation of
random placed and oriented Gabor elements that mask as much as possible the
aligned Gabor elements structure \cite{Demeyer11displays_using_GERT}.
Figure~\ref{fig:gabor-position} shows an example displaying only the elements
position; even if there is in fact a set of perfectly regularly aligned dots, it
is very hard to spot them. This suggests that the position of the elements
carries few useful cues about the alignment.

\subsection{The Detection Algorithm}

Let us now present the alignment detection algorithm that will be matched to
human perception. The input to the algorithm is a set of Gabor elements
$\mathbf{g}=\{(x_i,\theta_i)\}_{i=1\ldots N}$, defined by the position and
orientation of each element. We will further assume that the total number of
elements is a fixed quantity $N$.

\begin{figure}[b]
\centering
\includegraphics[width=0.35\textwidth]{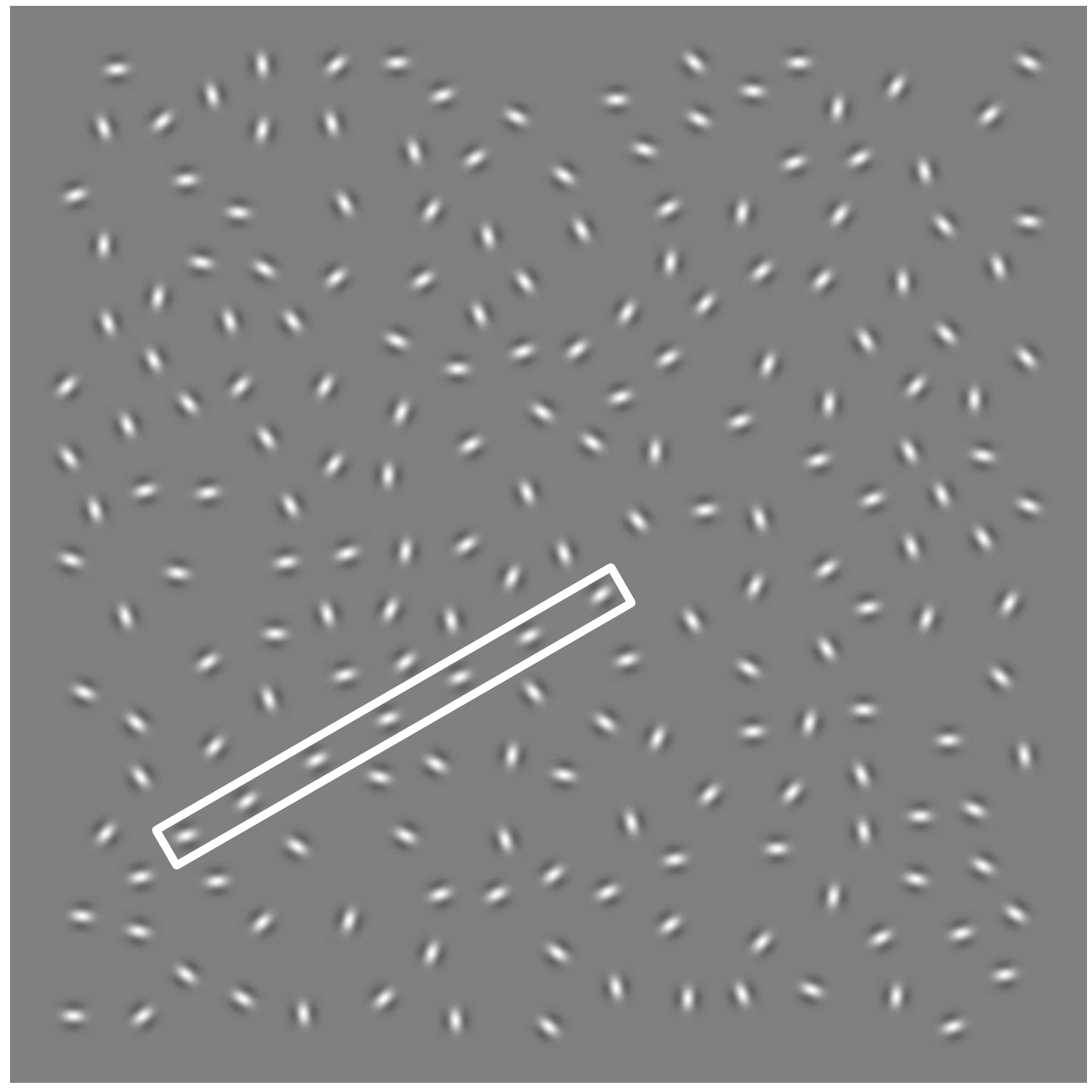}
\hfill
\begin{minipage}[b]{0.5\textwidth}
\centering
\includegraphics[width=0.55\textwidth]%
                {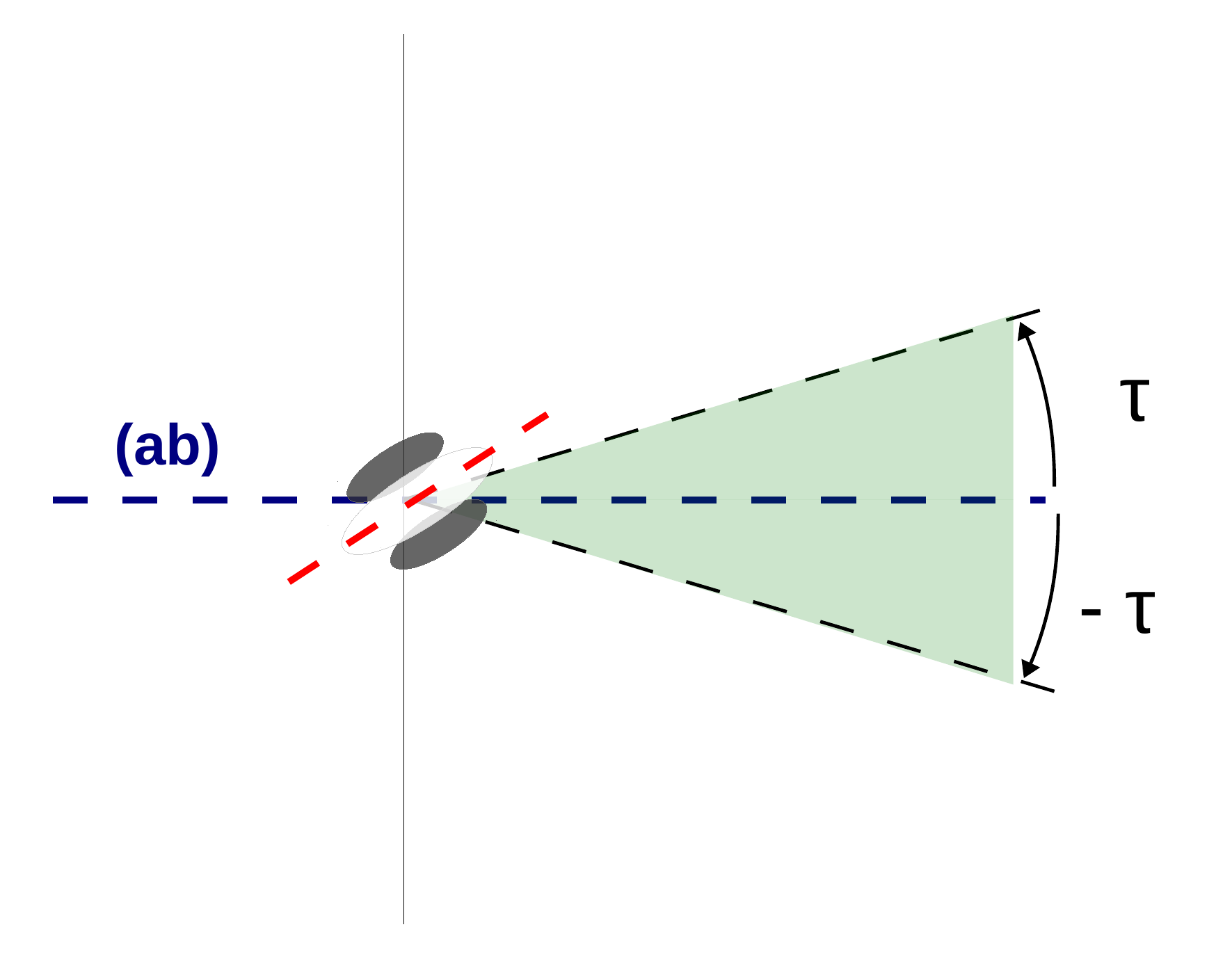}\\
\includegraphics[width=1\textwidth]{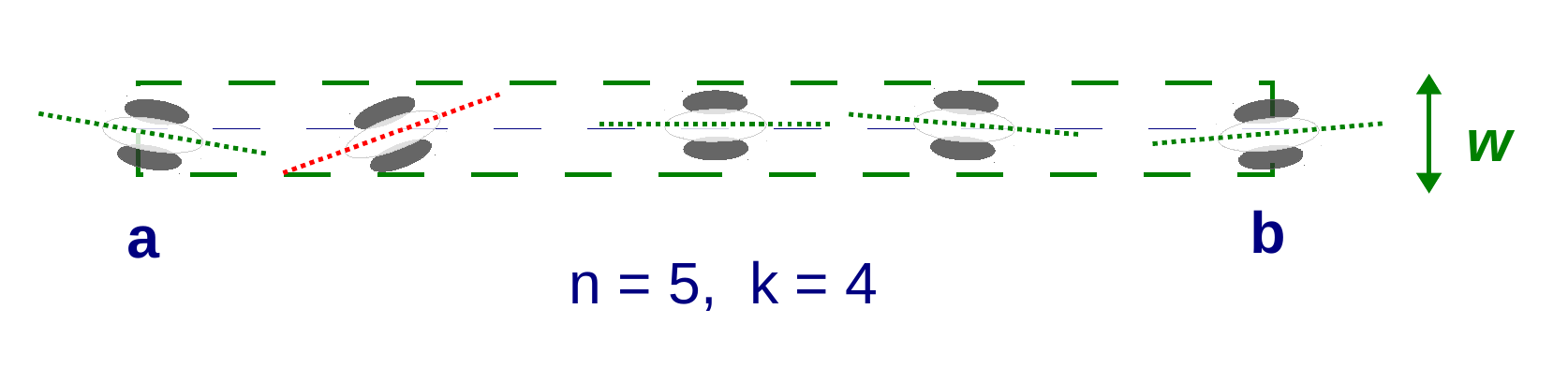}
\end{minipage}
\caption{\textbf{Left:} A candidate to alignment, defined by a rectangle
  $R$. \textbf{Right-Top:} A Gabor element whose angle with $(ab)$ is larger
  than $\tau$ and thus it is not counted as an aligned
  point. \textbf{Right-Bottom:} A detailed example where we see a total of five
  Gabor elements inside the rectangle, $n(r,\mathbf{g})=5$, being $\tau$-aligned
  with $(ab)$, i.e. $k_\tau(r,\mathbf{g})=4$.}
\label{fig:detection-algorithm}
\end{figure}

A candidate to alignment is defined as a rectangle $r$, see
Fig.~\ref{fig:detection-algorithm} (left), and the orientation of the Gabor
elements inside it will determine whether the candidate is evaluated as a valid
alignment or not. The orientation of each Gabor element is compared to the one
of the rectangle and when the difference is smaller than a given tolerance
threshold $\tau$, the element is said to be \emph{$\tau$-aligned}, see
Fig.~\ref{fig:detection-algorithm} (right). Two quantities will be observed for
each rectangle $r$: the total number of Gabor elements inside it,
$n(r,\mathbf{g})$, and the number among them that are $\tau$-aligned,
$k_{\tau}(r,\mathbf{g})$. The \emph{a contrario} validation is analogue to the
one described in Sect.~\ref{sec:basic_dot_align}.

Due to the way the patterns are generated, the only relevant information to
evaluate in an alignment is the orientation of the Gabor elements. Consequently,
the \emph{a contrario} model $H_0$ is defined with $N$ random variables
corresponding to the orientation of the elements and satisfying the following
two conditions:
\begin{itemize}
  \item the orientations $\Theta_i$ are independent from each other;
  \item each orientation $\Theta_i$ follows a uniform distribution in $[0,\pi)$.
\end{itemize}
Under these \emph{a contrario} assumptions, the probability that a Gabor element
be $\tau$-aligned to a given rectangle is given by
\begin{equation}
   p(\tau) = \frac{2\tau}{\pi}.
\end{equation}
Notice that the symmetric Gabor elements are unaltered by a rotation of $\pi$
rads. The independence hypothesis implies that the probability term
$\mathds{P}[k_{\tau}(r,\mathbf{G})\geq k_{\tau}(r,\mathbf{g})]$, where
$\mathbf{G}$ is a random set of Gabor elements following $H_0$, is given by
\begin{equation}
   \mathds{P}\big[k_{\tau} (r,\mathbf{G}) \geq k_{\tau}(r,\mathbf{g}) \big] =
         \mathcal{B}\Big(n(r,\mathbf{g}), k_{\tau} (r,\mathbf{g}), p(\tau) \Big),
\end{equation}
where as before $\mathcal{B}(n,k,p)$ is the tail of the binomial distribution.

We still need to specify the family of tests to be performed. Each pair of dots
will define a rectangle of fixed width $w$, so the total number of rectangles is
$\frac{N(N-1)}{2}$. Also, a finite number of angular precisions $\tau_i$ will be
tested for each rectangle. Then,
\begin{equation}
   N_{tests} = \frac{N(N-1)}{2} \cdot \#\mathcal{T},
\end{equation}
where $\#\mathcal{T}$ is the cardinality of the set $\mathcal{T}$ of
precisions. The NFA of a candidate is defined by
\begin{equation}
   \mathrm{NFA}(r,\mathbf{g}) = N_{tests} \cdot \min_{\tau\in\mathcal{T}}
          \mathcal{B}\Big(n(r,\mathbf{g}),k_{\tau}(r,\mathbf{g}), p(\tau)\Big).
\end{equation}
A large NFA value corresponds to a likely (and therefore insignificant)
configuration in the \emph{a contrario} model; inversely, a small NFA value
indicates a rare and interesting event. The proposed detection method validates
a rectangle candidate $r$ whenever $\mathrm{NFA}(r,\mathbf{x})<\varepsilon$. The
following theorem shows that it satisfies the non-accidentalness principle.
\begin{theorem}
$$
   \mathds{E}\left[\sum_{R\in\mathcal{R}} \mathds{1}_{\mathrm{NFA}(R,\mathbf{G})<\varepsilon}
             \right] \leq \varepsilon.
$$
where $\mathds{E}$ is the expectation operator, $\mathds{1}$ is the indicator
function, $\mathcal{R}$ is the set of rectangles considered, and $\mathbf{G}$ is
a random set of Gabor elements on $H_0$.
\end{theorem}

\begin{figure}[t]
\centering
\begin{tabular}{ccc}
\includegraphics[width=0.25\textwidth]{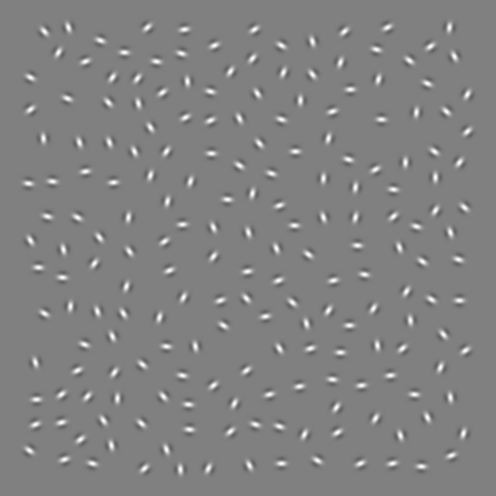} &
\includegraphics[width=0.251\textwidth]{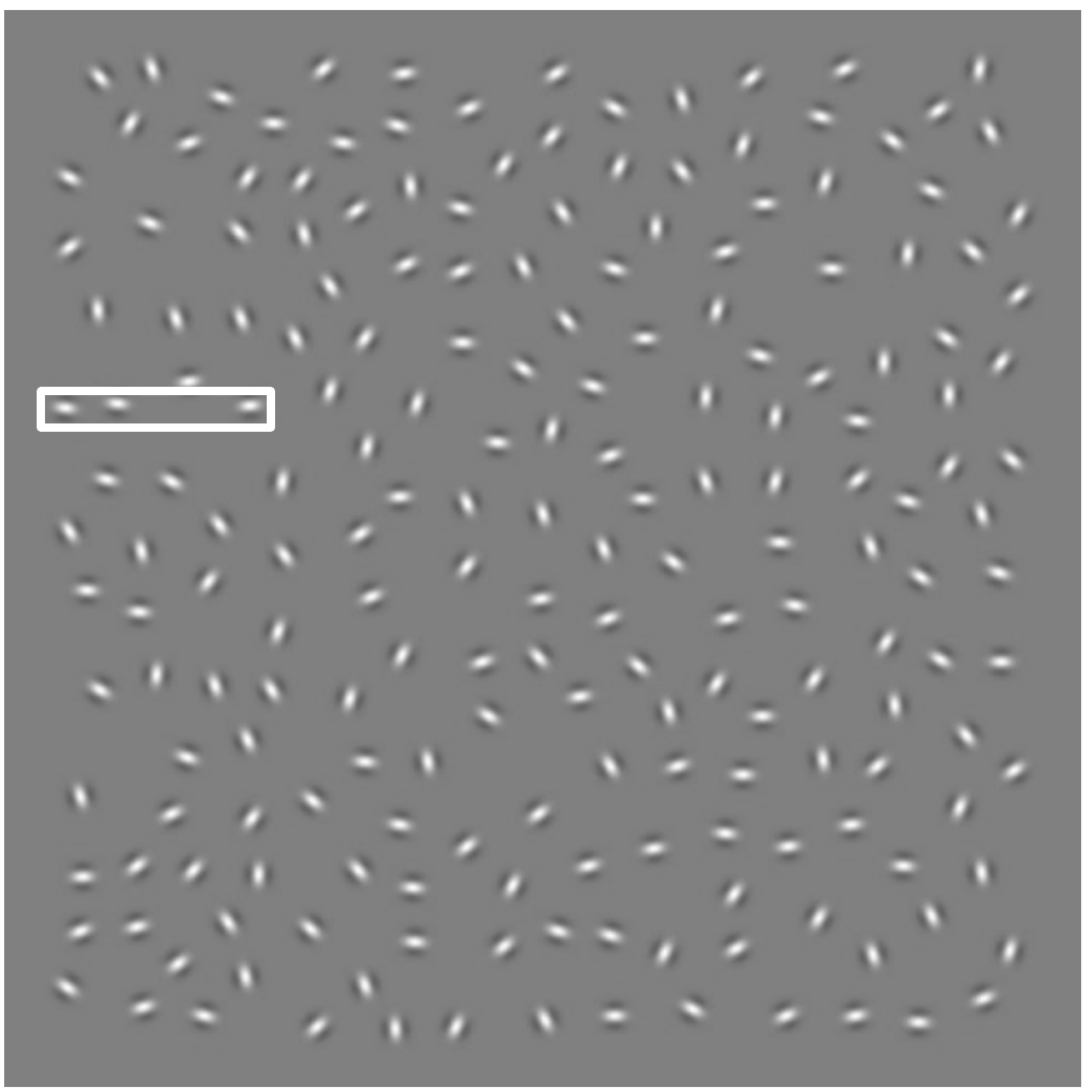} &
\includegraphics[width=0.25\textwidth]{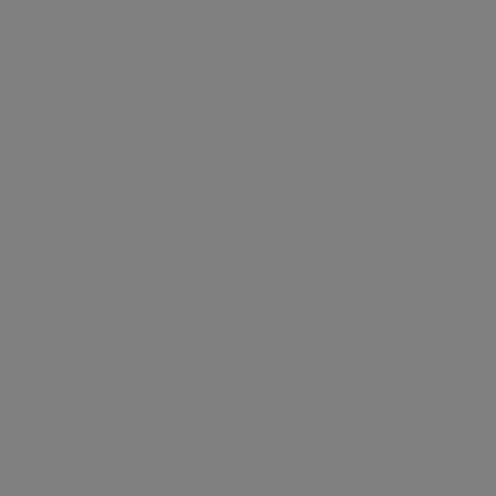} \\
& $\mathrm{NFA}=99.5$ & no detection \\
\includegraphics[width=0.25\textwidth]{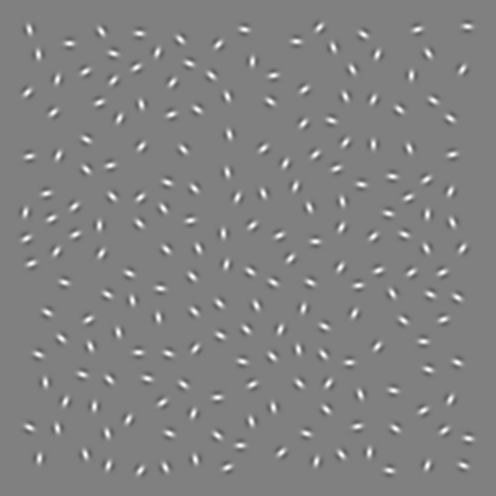} &
\includegraphics[width=0.251\textwidth]{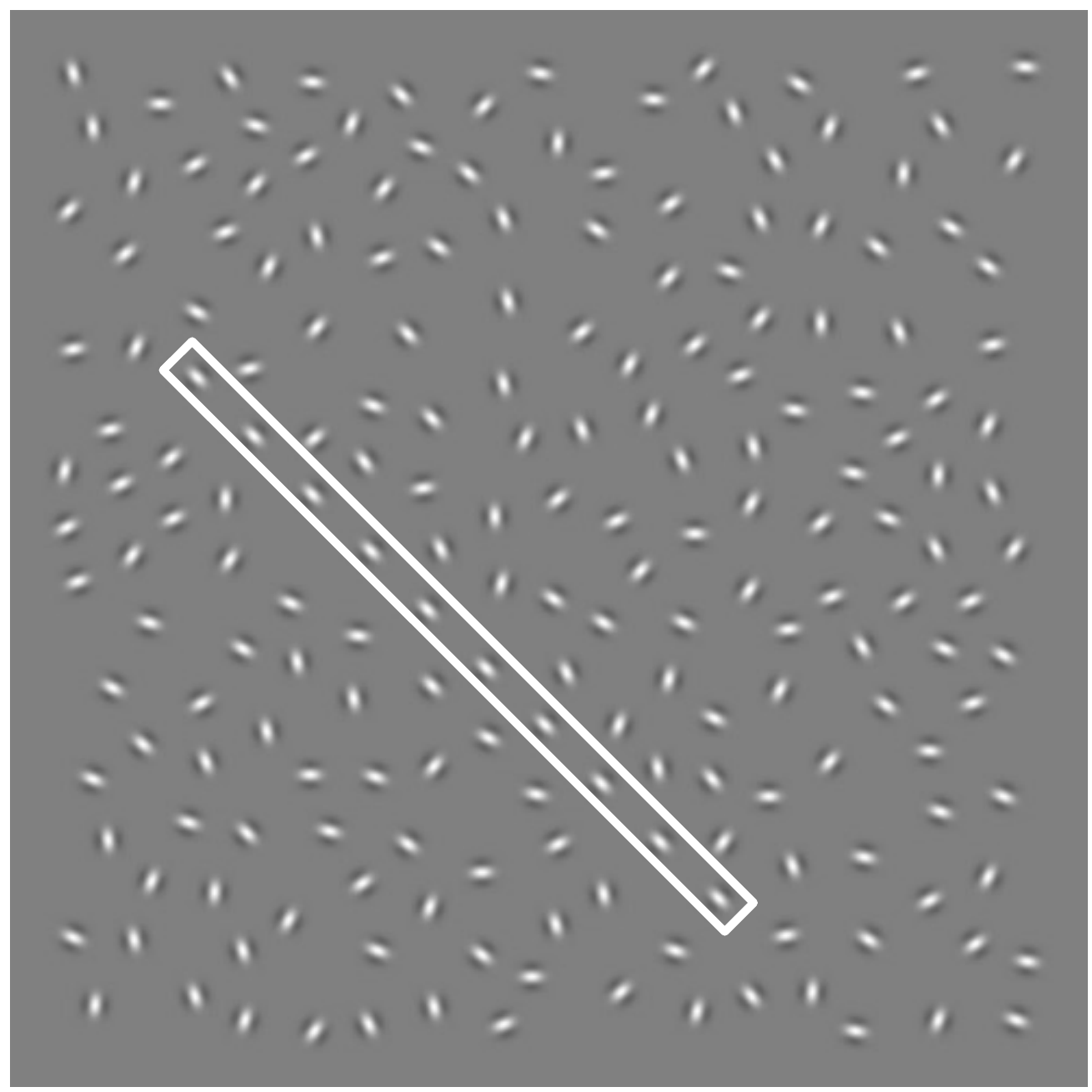} &
\includegraphics[width=0.25\textwidth]%
                {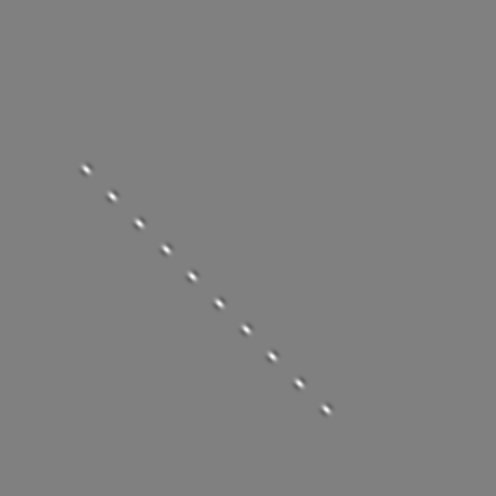} \\
& $\mathrm{NFA}=10^{-5}$ & detection \\
\end{tabular}
\caption{Two examples of the proposed validation method for alignment of Gabor
  elements. The rectangle in the first row has three elements inside, all of
  them aligned; that number is nevertheless too small to produce a detection, as
  its NFA value is larger than one. In the second example, all ten Gabor
  elements are aligned, giving an NFA of $10^{-5}$ and producing a detection.}
\label{fig:resultsAlgoLinesInBg}
\end{figure}

Once again we follow Desolneux~et~al. \cite{dmm2000,DMM_book} and set
$\varepsilon=1$. In our experiments, we use the NFA as an indication of the
visibility of the gestalt according to the proposed theory; a value considerably
smaller than 1 is ``non-accidental'' and should imply a conspicuous gestalt. A
value larger than 1 can occur just by chance and should therefore be associated
to an irrelevant gestalt. Figure~\ref{fig:resultsAlgoLinesInBg} shows two
examples of detection by this method.

\subsection{Experiment}

A psychophysical experiment was performed online by voluntary subjects using an
interactive web
site\footnote{\url{http://dev.ipol.im/~blusseau/aligned_gabors}}. Their task was
to report on the visibility of the aligned Gabor patterns. The online
methodology was necessarily more flexible and less controlled on various aspects
than it would be in a laboratory: we had no reliable information about the
subjects, their visualization conditions in front of their computers were not
controlled, the comprehension of the task by the subjects might vary,
etc. Notwithstanding their uncontrolled essence, online experiments give access
to a larger number of subjects and bring a great experimental flexibility.

The data set used for this experiment is composed of over 14000 stimuli
(negative and positive) as the one described in Sect.~\ref{sec:patterns}. Each
image has a size of $496 \times 496$ pixels and containing $N=200$ elements. For
positive stimuli, 9 levels of jitter ($J\in\lbrace 0, \frac{\pi}{5},
\frac{\pi}{4}, \frac{\pi}{3}, \frac{\pi}{2}, \frac{2\pi}{3}, \frac{3\pi}{4},
\frac{4\pi}{5}, \pi \rbrace$) and 8 different segment lengths were used, between
3 and 10 elements. During each session, the subject saw 35 of these images, one
after another. The first five images were training stimuli and no results were
recorded at this stage of the experiment. The following 30 images were randomly
sampled over the data set, with constraints that ensured a balance between
negative, positive, hard and easy stimuli. For each stimulus, the subject was
asked to answer whether they saw or not a ``straight line''; the answer and
response time were recorded. There was no time limit to provide the answer but
it was suggested to answer as soon as the subject made up their mind. At the end
of the session, a feedback was given on false detections and on the consistency
of the subject's answer through an ``attention score''. This score rewarded the
fact that the subject answered better on easy stimuli than on hard ones and
indicated if the task was well understood or not.

\subsection{Results}\label{sec:ResAnalysisAlgo}

In order to compare human and machine perception we precomputed the NFA for each
rectangle on all the images of the data set. Each image was associated to its
best (smallest) NFA. The hypothesis to be tested was that the NFA value should
be related directly to the visibility by humans; if this is true, the average
score given to an image by humans, namely the proportion of ``Yes'', should be
related to the NFA of the most salient structure. In what follows we will
analyze the data obtained from 7137 trials.

\begin{figure}[b]
\centering
\includegraphics[height=0.37\textwidth]{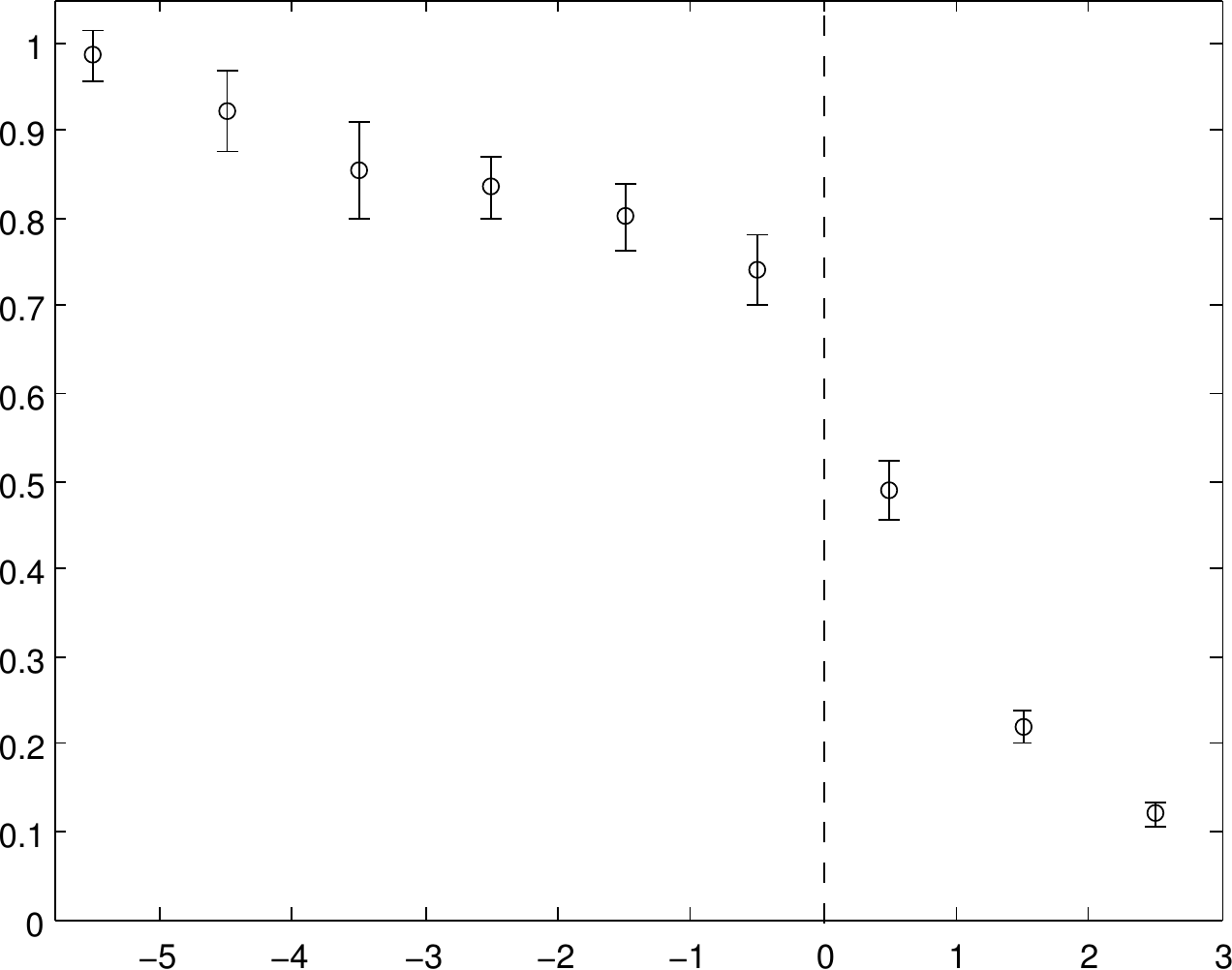}
\hfill
\includegraphics[height=0.37\textwidth]{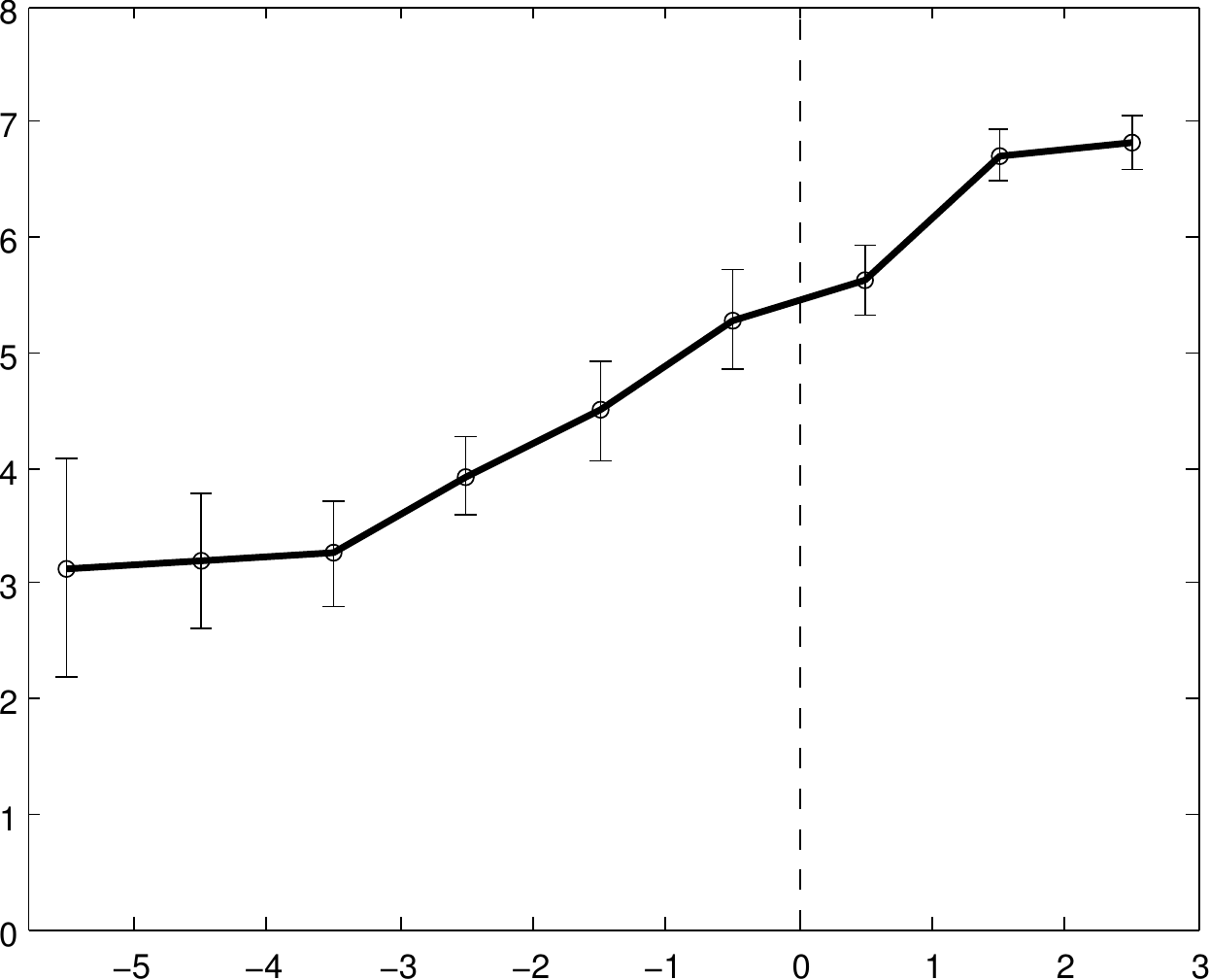}
\caption{Comparison of the subjects' responses to the NFA. \textbf{Left:} The
  average answer rate is plotted relative to $\log_{10}(\mathrm{NFA})$. Each
  point indicates the proportion of positive answers to stimuli with best NFA in
  the corresponding bin. \textbf{Right:} The average response times in seconds
  per bin. In both cases, the abscissa represents the scale of
  $\log_{10}(\mathrm{NFA})$ divided into 9 bins; the first bin is defined by
  $\log_{10}(\mathrm{NFA}) < -5$, the last one by $\log_{10}(\mathrm{NFA}) \geq
  2$, and the other 7 bins by $k \leq \log_{10}(\mathrm{NFA}) < k+1$ for $k= -5,
  \dots, 1 $. The error bars give approximately 95 \% confidence about the mean
  values (each interval is defined as $[\overline{x} - 2\frac{s}{\sqrt{n}},
    \overline{x} + 2\frac{s}{\sqrt{n}}]$, where $\overline{x}, s$ and $n$ are
  respectively the mean, standard deviation and number of trials of the bin).}
\label{fig:perf_time_VS_nfa}
\end{figure}

The NFA scale was divided into bins. To each bin were associated statistics on
the trials whose NFAs belonged to this bin. Figure~\ref{fig:perf_time_VS_nfa}
shows the average answer rate and response times for nine
$\log_{10}(\mathrm{NFA})$ intervals. Note that $\mathrm{NFA}<1$ (or
$\log_{10}(\mathrm{NFA})<0$) means detection of the alignment by the algorithm.

The results significantly support the hypothesis that a single scalar function
of all parameters predicts the detectability. Indeed, the answer rate follows a
sigmoid shape roughly centered at $\log_{10}(\mathrm{NFA})=0$. The second graph,
plotting the response time versus the NFA, also agrees with the hypothesis: the
less visible the stimuli are, the more time is spent searching for valid
gestalts. Statistical tests confirm this average tendency.

The experiment confirms the hypothesis that, at least in this restricted
perceptual environment (formed of three parameters, the number of Gabor patches,
the length of the alignment and its jitter on orientation), the value of NFA may
account for the human ``detectability'' of an alignment. Surprisingly, the human
detection (attentive) threshold is close to the best algorithm in this
restricted environment. Indeed, alignments with NFA smaller than 1 were detected
by a majority of subjects. Alignments with NFA larger than one, which are likely
to occur just by chance, were detected by a minority of subjects. Furthermore,
the detection curve is steepest when the NFA crosses 1. The curve is not as
steep for the mean response time as a function of NFA. This can be simply
explained by the fact that the patience of subjects undergoes a rapid temporal
erosion; they are not ready to look long for a needle in a haystack.

\subsection{Consequence: An Online Game}

Online experimentation opens new possibilities that need to be explored farther,
and in particular the use of computer games as an experimentation tool. A
successful game may attract the attention of subjects and if the resulting mass
of results is large enough, it could compensate for the lack of control on other
aspects of the experimental setting.

The player of a computer game is usually directed toward an objective and faced
with obstacles. To be attractive, a game cannot be too easy, but not too hard
either; a good balance of this difficulty is the key to the popularity of the
game. To use games for psychophysical purposes, the player should be directed to
detect some pattern, the obstacles being the conditions that preclude this
perception. Motivated players will do their best effort, revealing the limits of
human perception.

To go in this direction we created a prototype version of an alignment
game.\footnote{\url{http://dev.ipol.im/~blusseau/clickline}} The player is
presented with Gabor stimuli as described before. Only positive stimuli with
variable difficulties are used. In this way one knows that there is an alignment
gestalt; but its position is unknown and the assignment of the player is to spot
it. The subject is asked to click in the image on any point of the straight
line. The distance between the clicked point and the actual line segment is
recorded and a score over 100 is computed as a function of this distance (the
closer to the segment, the better the score). When the stimulus is quite
visible, all subjects are able to point correctly to it; when it is not, the
distance to the alignment becomes random. This rash transition should permit to
pinpoint the human detection threshold.

The presentation of the stimuli is divided into several sequences of ten
images. The first sequence is always supposed to be very easy (long segments
with little jitter). Then the difficulty of the following sequences change
according to the performance achieved on the previous one. The collected data
will permit us to compare a detection method to human performance in the way
described before. The game is still in a prototype phase but readers are invited
to try it and provide feedback.

\section{Conclusion}

Needless to be said, the experimental devices and first results that we just
described are not sufficient to make any rash conclusion on the existence of
quantitative predictions of human perception. They will need to be extended to
other gestalts commonly used in psychophysics, such as for example contours
(good continuation), clusters, or symmetries. In the same way, the first
described gestaltic game does not furnish an end algorithm modeling what we
could call the human notion of alignment. Finally, we did not deliver a
detection algorithm directly usable on any image, as required by the computer
vision methodology. In short, this is work in progress, and our goal was to
raise the attention of psychophysical researchers and computer scientists on the
interest of introducing Turing tests in their methodology.

\bibliographystyle{spmpsci}
\bibliography{biblio}

\begin{thebibliography}{10}
\providecommand{\url}[1]{{#1}}
\providecommand{\urlprefix}{URL }
\expandafter\ifx\csname urlstyle\endcsname\relax
  \providecommand{\doi}[1]{DOI~\discretionary{}{}{}#1}\else
  \providecommand{\doi}{DOI~\discretionary{}{}{}\begingroup
  \urlstyle{rm}\Url}\fi

\bibitem{Ahuja89}
Ahuja, N., Tuceryan, M.: Extraction of early perceptual structure in dot
  patterns: integrating region, boundary, and component gestalt.
\newblock Comput. Vision Graph. Image Process. \textbf{48}(3), 304--356 (1989)

\bibitem{Attneave54InformationalAspects}
Attneave, F.: Some informational aspects of visual perception.
\newblock Psychological Review \textbf{61}(3), 183--193 (1954)

\bibitem{Demeyer11displays_using_GERT}
Demeyer, M., Machilsen, B.: The construction of perceptual grouping displays
  using \uppercase{gert}.
\newblock Behavior Research Methods, online first. pp. 1--8 (2011)

\bibitem{dmm2000}
Desolneux, A., Moisan, L., Morel, J.: Meaningful alignments.
\newblock International Journal of Computer Vision \textbf{40}(1), 7--23 (2000)

\bibitem{es2003computational}
Desolneux, A., Moisan, L., Morel, J.: Computational gestalts and perception
  thresholds.
\newblock Journal of Physiology - Paris \textbf{97}, 311--324 (2003)

\bibitem{DMM2003}
Desolneux, A., Moisan, L., Morel, J.: A grouping principle and four
  applications.
\newblock IEEE Transactions on Pattern Analysis and Machine Intelligence
  (2003)

\bibitem{DMM_book}
Desolneux, A., Moisan, L., Morel, J.: From {G}estalt Theory to Image Analysis,
  a Probabilistic Approach, \emph{Interdisciplinary Applied Mathematics},
  vol.~34.
\newblock Springer (2008)

\bibitem{source-book}
Ellis, W. (ed.): A Source Book of {G}estalt Psychology.
\newblock Humanities Press (1967 (originally 1938))

\bibitem{feldman1997regularity}
Feldman, J.: Regularity-based perceptual grouping.
\newblock Computational Intelligence \textbf{13}(4), 582--623 (1997)

\bibitem{Feldman01BayesianContour}
Feldman, J.: Bayesian contour integration.
\newblock Attention, Perception, \& Psychophysics \textbf{63}, 1171--1182
  (2001)

\bibitem{Feldman05InfoAlongContours}
Feldman, J., Singh, M.: Information along contours and object boundaries.
\newblock Psychological Review \textbf{112}(1), 243--252 (2005)

\bibitem{Field93ContourIntegration}
Field, D.J., Hayes, A., Hess, R.F.: Contour integration by the human visual
  system: Evidence for a local “association field”.
\newblock Vision Research \textbf{33}(2), 173 -- 193 (1993)

\bibitem{geman-etal-pnas}
Fleuret, F., Li, T., Dubout, C., Wampler, E.K., Yantis, S., Geman, D.:
  Comparing machines and humans on a visual categorization test.
\newblock Proceedings of the National Academy of Sciences \textbf{108}(43),
  17,621--17,625 (2011)

\bibitem{GJ08}
Grompone~von Gioi, R., Jakubowicz, J.: On computational {G}estalt detection
  thresholds.
\newblock Journal of Physiology -- Paris \textbf{103}(1-2), 4--17 (2009)

\bibitem{grossberg1985neural}
Grossberg, S., Mingolla, E.: Neural dynamics of perceptual grouping: Textures,
  boundaries, and emergent segmentations.
\newblock Attention, Perception, \& Psychophysics \textbf{38}(2), 141--171
  (1985)

\bibitem{han-zhu}
Han, F., Zhu, S.C.: Bottom-up/top-down image parsing with attribute grammar.
\newblock IEEE Transactions on Pattern Analysis and Machine Intelligence
  \textbf{31}(1), 59--73 (2009)

\bibitem{kanizsa1979organization}
Kanizsa, G.: Organization in vision: Essays on Gestalt perception.
\newblock Praeger New York: (1979)

\bibitem{kanizsa}
Kanizsa, G.: Grammatica del vedere.
\newblock Il Mulino (1980)

\bibitem{kanizsa:vedere}
Kanizsa, G.: Vedere e pensare.
\newblock Il Mulino (1991)

\bibitem{Kersten04ObjectPerceptionBayesianInference}
Kersten, D., Mamassian, P., Yuille, A.: Object perception as bayesian
  inference.
\newblock Annual Review of Psychology \textbf{55}(1), 271--304 (2004)

\bibitem{kohler}
K{\"o}hler, W.: Gestalt Psychology.
\newblock Liveright (1947)

\bibitem{leclerc1989constructing}
Leclerc, Y.: Constructing simple stable descriptions for image partitioning.
\newblock International journal of computer vision \textbf{3}(1), 73--102
  (1989)

\bibitem{Lowe85}
Lowe, D.: Perceptual Organization and Visual Recognition.
\newblock Kluwer Academic Publishers (1985)

\bibitem{marr}
Marr, D.: Vision.
\newblock Freeman and co. (1982)

\bibitem{metzger}
Metzger, W.: Gesetze des Sehens, third edn.
\newblock Verlag Waldemar Kramer, Frankfurt am Main (1975)

\bibitem{metzger:en}
Metzger, W.: Laws of Seeing.
\newblock The MIT Press (2006 (originally 1936)).
\newblock English translation of the first edition of \cite{metzger}.

\bibitem{mumford}
Mumford, D.: Pattern theory: the mathematics of perception.
\newblock Proceedings of the International Congress of Mathematicians, Beijing
  \textbf{I}, 401--422 (2002)

\bibitem{Nygard09orientation_jitter_motion}
Nyg{\aa}rd, G., Van~Looy, T., Wagemans, J.: The influence of orientation jitter
  and motion on contour saliency and object identification.
\newblock Vision Research \textbf{49}, 2475--2484 (2009)

\bibitem{pinar2000turing}
Pinar~Saygin, A., Cicekli, I., Akman, V.: Turing test: 50 years later.
\newblock Minds and Machines \textbf{10}(4), 463--518 (2000)

\bibitem{PreissThesis}
Preiss, K.: A theoretical and computational investigation into aspects of human
  visual perception: Proximity and transformations in pattern detection and
  discrimination.
\newblock Ph.D. thesis, University of Adelaide (2006)

\bibitem{sarkar1993}
Sarkar, S., Boyer, K.L.: Perceptual organization in computer vision: A review
  and a proposal for a classificatory structure.
\newblock IEEE Transactions on Systems, Man, and Cybernetics \textbf{23}(2),
  382--399 (1993)

\bibitem{spelke1990principles}
Spelke, E.: Principles of object perception.
\newblock Cognitive science \textbf{14}(1), 29--56 (1990)

\bibitem{stevens}
Stevens, S.: Psychophysics.
\newblock Transaction Publishers (1986)

\bibitem{Tripathy99}
Tripathy, S.P., Mussap, A.J., Barlow, H.B.: Detecting collinear dots in noise.
\newblock Vision Research \textbf{39}(25), 4161 -- 4171 (1999)

\bibitem{turing1950}
Turing, A.: Computing machinery and intelligence.
\newblock Mind \textbf{59}, 433--460 (1950)

\bibitem{Uttal70}
Uttal, W., Bunnell, L., Corwin, S.: On the detectability of straight lines in
  visual noise: An extension of french’s paradigm into the millisecond
  domain.
\newblock Perception and Psychophysics \textbf{8}, 385--388 (1970)

\bibitem{Uttal73}
Uttal, W.R.: The effect of deviations from linearity on the detection of dotted
  line patterns.
\newblock Vision Res \textbf{13}(11), 2155--63 (1973)

\bibitem{Vanegas10}
Vanegas, M.C., Bloch, I., Inglada, J.: Detection of aligned objects for high
  resolution image understanding.
\newblock In: IGARSS, pp. 464--467 (2010)

\bibitem{wagemans-nonaccidental}
Wagemans, J.: Perceptual use of nonaccidental properties.
\newblock Canadian Journal of Psychology \textbf{46}(2), 236--279 (1992)

\bibitem{Wagemans12CenturyOfGestalt-I}
Wagemans, J., Elder, J.H., Kubovy, M., Palmer, S.E., Peterson, M.A., Singh, M.,
  von~der Heydt, R.: A century of \uppercase{G}estalt psychology in visual
  perception: {I}. \uppercase{P}erceptual grouping and figure--ground
  organization.
\newblock Psychological bulletin \textbf{138}(6), 1172--1217 (2012).
\newblock \doi{10.1037/a0029333}

\bibitem{Wageman12CenturyOfgestalt-II}
Wagemans, J., Feldman, J., Gepshtein, S., Kimchi, R., Pomerantz, J.R., van~der
  Helm, P.A., van Leeuwen, C.: A century of {G}estalt psychology in visual
  perception: {II}. {C}onceptual and theoretical foundations.
\newblock Psychological Bulletin  (2012)

\bibitem{wertheimer}
Wertheimer, M.: {U}ntersuchungen zur {L}ehre von der {G}estalt. {II}.
\newblock Psychologische Forschung \textbf{4}(1), 301--350 (1923).
\newblock An abridged translation to English is included in \cite{source-book}.

\bibitem{witkin2}
Witkin, A.P., Tenenbaum, J.M.: On the role of structure in vision.
\newblock In: J.~Beck, B.~Hope, A.~Rosenfeld (eds.) Human and {M}achine
  {V}ision, pp. 481--543. Academic Press (1983)

\bibitem{witkin1}
Witkin, A.P., Tenenbaum, J.M.: What is perceptual organization for?
\newblock IJCAI-83 \textbf{2}, 1023--1026 (1983)

\bibitem{zhu1996region}
Zhu, S., Yuille, A.: Region competition: Unifying snakes, region growing, and
  bayes/mdl for multiband image segmentation.
\newblock Pattern Analysis and Machine Intelligence, IEEE Transactions on
  \textbf{18}(9), 884--900 (1996)

\bibitem{zhu-mumford}
Zhu, S.C., Mumford, D.: A stochastic grammar of images.
\newblock Foundations and Trends in Computer Graphics and Vision \textbf{2}(4),
  259--362 (2006)

\end{thebibliography}

\end{document}